\documentclass[preprint,12pt]{elsarticle}

\UseRawInputEncoding   
\usepackage{textcomp}    
\usepackage{amsmath}     
\usepackage{amssymb}     
\usepackage{amsfonts}    
\usepackage{amsthm}      
\usepackage{graphicx}
\usepackage{subcaption} 
\usepackage{placeins}   
\usepackage{booktabs}    
\usepackage{multirow}   
\usepackage{multicol}   
\usepackage{tabularx}
\usepackage{rotating}
\usepackage{makecell}    
\usepackage{array}       
\usepackage{threeparttable} 
\usepackage{tablefootnote}  
\usepackage[table,xcdraw]{xcolor} 
\usepackage[normalem]{ulem} 
\usepackage{algorithm}
\usepackage{algorithmic}
\usepackage{url}        
\usepackage{verbatim}    
\usepackage{ragged2e}    
\biboptions{numbers,sort&compress}
\usepackage[hidelinks]{hyperref}
\newcommand{\stat}{\textsuperscript{***}}

\journal{**}

\begin{document}

\begin{frontmatter}

\title{Frequency Error-Guided Under-sampling Optimization for Multi-Contrast MRI Reconstruction}

\author[label1]{Xinming Fang}
\ead{fangxinming@shu.edu.cn}

\author[label2]{Chaoyan Huang}
\ead{huang345@msu.edu}

\author[label3]{Juncheng Li\corref{cor1}}
\ead{jcli@cs.ecnu.edu.cn}

\author[label1]{Jun Wang}
\ead{wangjun_shu@shu.edu.cn}

\author[label1]{Jun Shi}
\ead{junshi@shu.edu.cn}

\author[label3]{Guixu Zhang}
\ead{gxzhang@cs.ecnu.edu.cn}

\cortext[cor1]{Corresponding author.}
\cortext[cor2]{Equal contribution: Xinming Fang, Chaoyan Huang}

\address[label1]{School of Communication and Information Engineering, Shanghai University, Shanghai, 200444, China}
\address[label2]{School of Computational Mathematics Science and Engineering, Michigan State University, Ann Arbor, USA}
\address[label3]{School of Computer Science and Technology, East China Normal University, Shanghai, 200062, China}

\begin{abstract}
Magnetic resonance imaging (MRI) plays a vital role in clinical diagnostics, yet it remains hindered by long acquisition times and motion artifacts. Multi-contrast MRI reconstruction has emerged as a promising direction by leveraging complementary information from fully-sampled reference scans. However, existing approaches suffer from three major limitations: (1) superficial reference fusion strategies, such as simple concatenation, (2) insufficient utilization of the complementary information provided by the reference contrast, (3) fixed under-sampling patterns.
We propose an efficient and interpretable frequency error-guided reconstruction framework to tackle these issues.
We first employ a conditional diffusion model to learn a Frequency Error Prior (FEP), which is then incorporated into a unified framework for jointly optimizing both the under-sampling pattern and the reconstruction network. 
The proposed reconstruction model employs a model-driven deep unfolding framework that jointly exploits frequency- and image-domain information. In addition, a spatial alignment module and a reference feature decomposition strategy are incorporated to improve reconstruction quality and bridge model-based optimization with data-driven learning for improved physical interpretability.
Comprehensive validation across multiple imaging modalities, acceleration rates (4-30$\times$), and sampling schemes demonstrates consistent superiority over state-of-the-art methods in both quantitative metrics and visual quality. All codes are available at \url{https://github.com/fangxinming/JUF-MRI}.
\end{abstract}


\end{frontmatter}

\section{Introduction}
\label{sec:introduction}
Magnetic resonance imaging (MRI) is one of the most versatile medical imaging modalities, offering non-ionizing acquisition, superior soft-tissue contrast, and high spatial resolution. In clinical practice, multiple contrasts (e.g., T1-weighted, T2-weighted, PD-weighted) are often acquired to enhance diagnostic accuracy. However, each contrast requires fully sampled $k$-space data, and certain scans demand longer echo time and repetition time \cite{zhou2020dudornet}. Extended scan durations not only increase patient discomfort and the likelihood of motion artifacts but may also delay critical treatment decisions~\cite{feng2022multimodal}. Therefore, seeking an accelerated MRI reconstruction method is important for reducing acquisition time and enhancing image quality. Traditional techniques include compressed sensing (CS) \cite{yang2018admm} and parallel imaging (PI) \cite{pieciak2016non}. CS-MRI enforces sparsity in transform domains through regularized iterative solvers~\cite{lustig2007sparse,liang2009accelerating}, but at high acceleration ratios it can suffer residual aliasing and noise~\cite{ravishankar2010mr}. PI exploits multi-coil sensitivity profiles to compensate missing data~\cite{deshmane2012parallel}, yet experiences increased g-factor penalties and artifacts when in-plane acceleration exceeds two~\cite{hutchinson1988fast,feng2021multi}.

Recently, numerous deep learning-based MRI reconstruction methods have achieved superior reconstruction results. These methods can be broadly categorized into single-contrast and multi-contrast reconstruction. Single-contrast methods focus on reconstructing MR images from under-sampled $k$-space data of a single imaging contrast~\cite{wang2016accelerating,jin2017deep,dedmari2018complex,lee2018deep,meng2025dh,sun2025fourier,shin2025ensemble}, typically relying solely on information within that specific contrast. On the other hand, multi-contrast methods utilize images from different contrasts as reference to help reconstruct the target contrast image \cite{zhou2020dudornet,feng2022multimodal,feng2021multi,xiang2018deep,xiang2018ultra,zou2025mmr}. Due to the substantial structural redundancy across different contrasts from the same anatomical region, the reference images provide valuable complementary information, enabling more accurate reconstruction and mitigating the generation of non-existent anatomical features~\cite{lyu2020multi,li2022transformer,zhou2020dudornet}. Furthermore, the acquisition durations of different MRI contrasts vary considerably; for instance, T1-weighted images typically require less time than T2-weighted images~\cite{zhou2020dudornet,lei2024joint}. By using faster-acquired contrasts as reference modalities, it becomes possible to improve the reconstruction of slower-acquired under-sampled contrasts while keeping the total acquisition time within a clinically acceptable range.

Following this line, numerous multi-contrast MRI reconstruction (MC-MRI) schemes have been proposed. For example, ~\cite{liu2021regularization,xiang2018deep,xiang2018ultra} concatenate the reference and the target contrast image along the channel dimension as the input for MRI reconstruction. However, such a straightforward fusion approach ignores the cross-contrast structural correlations and complementary semantic information inherent in multi-contrast data \cite{lyu2025fast}. Besides, the strategy of using predefined under-sampling masks to simulate accelerated acquisitions was applied in \cite{zhou2020dudornet,feng2022multimodal}. This sampling limits the ability to fully exploit the available information in $k$-space. Given that MR images are reconstructed via the inverse Fourier transform of $k$-space data, incorporating frequency-domain information into the learning process can improve reconstruction fidelity. Moreover, purely data-driven deep learning models for MRI reconstruction rely on manually designed architectures without incorporating domain-specific physical constraints, thereby lacking physical interpretability and consistency with MRI acquisition principles. 

\begin{figure*}[htbp]
    \centering
    \includegraphics[width=1\textwidth]{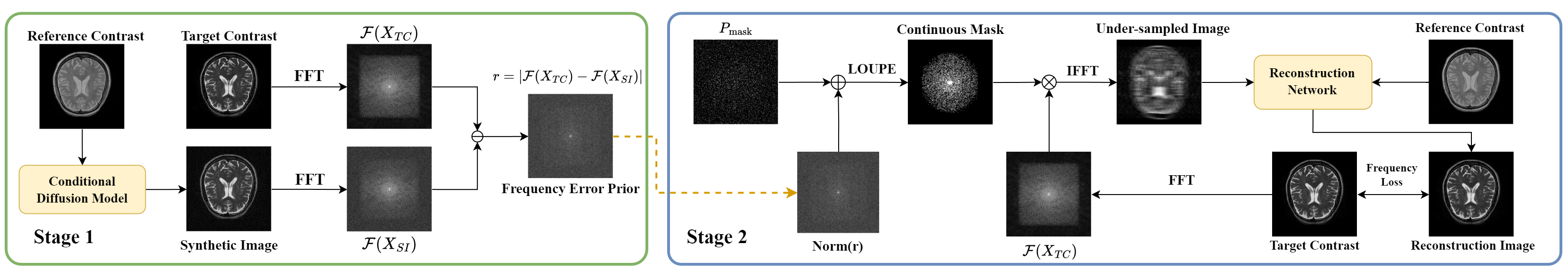}
    \caption{An illustration of the proposed JUF-MRI framework, which consists of two stages. In Stage 1, a frequency error prior $r$ is obtained using a conditional diffusion model. In Stage 2, the under-sampling pattern and the reconstruction network parameters are jointly optimized.}
    \label{fig:overall_structure}
\end{figure*}

To address limitations such as inefficient reference modality utilization, fixed undersampling patterns, and poor physical interpretability, we propose JUF-MRI (Joint Under-sampling optimization with Frequency error for MC-MRI), a new deep unfolding framework illustrated in Fig.~\ref{fig:overall_structure}. The framework operates in two stages. First, a conditional diffusion model synthesizes the target modality from the reference. Then, we apply a Fast Fourier Transform (FFT) to both the synthesized and target images, and their discrepancy in the frequency domain defines the frequency error $r$. This error captures critical modality-specific information that standard diffusion-based synthesis often misses. In the second stage, inspired by the LOUPE~\cite{bahadir2020deep} method, we adapt it to MC-MRI by incorporating $r$ as prior knowledge. This facilitates the joint optimization of the undersampling mask and the reconstruction network, guided by the frequency error.
As the reconstruction network employs a model-driven deep unfolding architecture, integrating $r$ can inject data-driven knowledge, thereby integrating the strengths of both model-driven and data-driven approaches. Meanwhile, we design a new reconstruction network based on the deep unfolding framework, wherein $k$-space data is explicitly integrated into the objective function to enrich the information utilized during training.
Unlike purely data-driven MC-MRI methods, the proposed model improves physical interpretability by coupling model-based optimization with data-driven learning in a deep unfolding framework. The main contributions can be summarized as follows: 

\begin{itemize}
\item We propose a new deep unfolding framework for MC-MRI, named JUF-MRI, which jointly under-sampling optimization with frequency error. This framework can solve the problems of insufficient reference modality utilization, fixed under-sampling patterns, and insufficient $k$-space data integration.

\item We propose a novel approach that employs a conditional diffusion model to obtain the frequency error $r$ between reference and target modalities, which is subsequently integrated as guiding prior for the reconstruction process.

\item We propose a new frequency error guided Reconstruction Network based on the deep unfolding framework, wherein frequency-domain information is explicitly integrated into the objective function to enrich the information utilized during training. Meanwhile, a spatial alignment module is introduced to mitigate spatial misalignment between reference and target modalities.
\end{itemize}

\section{Related works}
\subsection{Traditional Methods for MC-MRI}
Multi-contrast MRI reconstruction is usually divided into two steps. Firstly, model the problem with a mathematical function. Secondly, the designed algorithm is used to find the optimal solution. Specifically, the acquisition process can be roughly modeled as 
\begin{equation}\label{equation1}
    \widetilde{K}= M\mathcal{F}(X)+ \epsilon,
\end{equation}
where ${X} \in \mathbb{R}^{m\times n}$ is the latent fully sampled MR image, $\widetilde{K} \in \mathbb{C}^{m\times n}$ represents the observed $k$-space data, $M$ is the binary under-sampling mask, $\mathcal{F}$ denotes the Fourier transform, $\epsilon$ represents noise during the sampling. The goal is to find an appropriate solution $\hat{X}$ from the ill-posed model 
\begin{equation}\label{squarenorm}
    \hat{X}=\arg\min_{X}\frac{1}{2}\|M\mathcal{F}(X)-\widetilde{K}\|_{F}^{2}
\end{equation}
with designated methods. Naturally, the prior information is added to better find the $\hat{X}$ in the well-posed model 
\begin{equation}
\hat{X}=\arg\min_{X}\frac{1}{2}\|M\mathcal{F}(X)-\widetilde{K}\|_{F}^{2}+\lambda R(X,Y),
\label{equation3}
\end{equation}
where $R(X, Y)$ is the regularization, which incorporates prior knowledge between the reference image and the target image, and $\lambda$ is a positive trade-off parameter. 

Based on Eq.~\eqref{equation3}, manually designed regularizations are widely used. For example, the total variation (TV) and wavelet regularization terms are used in~\cite{bilgic2011multi,huang2014fast} to better model data sparsity and edge information. 
However, directly applying classical variational regularizations for MC-MRI usually results in unsatisfactory results. To improve the reconstruction quality, a spatial misalignment between the reference and target images was suggested in~\cite{DU2012compressed}. 

Most traditional methods use proven regularization methods to reconstruct MC-MRI images and achieve reasonable results using well-known algorithms such as the alternating direction method of multipliers (ADMM). However, these methods are not sufficient to cope with complex imaging tasks due to the neglect of algorithm adaptability, limited data utilization, and weak generalizability.

\subsection{Data-driven Methods for MC-MRI}
The core of data-driven deep learning methods is to train a deep neural network on large-scale datasets. For MC-MRI, the goal is to learn the mapping between the under-sampled MR image $X_u$ and the high-quality image $X$ with the support of the reference image $Y$. Generally, it can be modeled as 
\begin{equation}\label{equation5}
    X=\varphi_\theta ( X_{u}, Y), 
\end{equation}
where $\varphi_{\theta}$ is the learned neural network with parameter $\theta$ and ${X}_{u}=\mathcal{F}^{-1} (\tilde{K})$ with the inverse Fourier transform $\mathcal{F}^{-1}$.

Based on Eq.~\eqref{equation5}, some methods have been proposed for MC-MRI. For example, Do et al.~\cite{do2020reconstruction} modified the UNet to the Y-Net version, which separated the reference modality and the target modality into two input paths. 
Zhou et al.~\cite{zhou2020dudornet} proposed a DuDoRNet, which equipped the residual learning, recursive learning, dense connections, and dilated convolution for fast MRI reconstruction. The strategy of using two independent convolutional recurrent neural networks was proposed in~\cite{wang2020improving} to encourage a bi-directional flow among image features.

These data-driven deep learning methods produce high-quality reconstructed MRI images and achieve fast reconstruction speeds once the network training is complete. However, the learning strategies and module designs in these networks are usually manually crafted, lacking interpretability.

\subsection{Model-driven Methods for MC-MRI}
To ensure the safety and transparency of medical image reconstruction, recent advances have integrated mathematical modeling into deep learning frameworks, forming a new model-driven deep learning approach. A representative and effective strategy in this paradigm is deep unfolding, which embeds classic iterative optimization algorithms into neural network architectures.

A typical example of this approach is the model-driven deep attention network (MD-DAN) proposed by Yang et al.~\cite{yang2020model}, where the half-quadratic splitting (HQS) algorithm was unrolled to solve the MC-MRI reconstruction problem. Although MD-DAN effectively addresses prior modeling, it does not explicitly account for variations in image structure and contrast across different modalities. To overcome these issues, Lei et al.~\cite{lei2023decomposition} proposed a multi-contrast variational model that decomposes the reference image into consistent components and inconsistent components. 
Sun et al.~\cite{sun2023joint} observed that many reconstruction methods overlook the role of coil sensitivity estimation, which is crucial in multi-coil MRI. They proposed a joint optimization framework that unrolls the reconstruction process alongside iterative estimation of coil sensitivities, improving overall reconstruction accuracy and mitigating common artifacts.

Most existing model-driven methods use a fixed under-sampling pattern, usually designed heuristically or empirically. This ignores the potential interaction between sampling strategies and reconstruction networks. Therefore, even if the best reconstruction model is designed, poor performance may still occur when paired with sampling masks with poor matching.

\subsection{Under-sampling Pattern Optimization Methods}
A key step in MRI reconstruction is the choice of under-sampling strategy. At the same acceleration rate, different under-sampling methods will lead to different reconstruction quality. Therefore, learning-based under-sampling optimization has become a growing trend. The goal is to jointly optimize the sampling pattern and the reconstruction model in an end-to-end fashion. For example, Bahadir et al.~\cite{bahadir2020deep} proposed LOUPE, which jointly learns the under-sampling pattern and reconstruction network parameters, enabling adaptive mask design that improves reconstruction performance. 
Sherry et al.~\cite{sherry2020learning} proposed a bilevel supervised learning method that jointly learns sparse sampling patterns and variational reconstruction to accelerate MRI acquisition. Lei et al.~\cite{lei2024joint} combined LOUPE with an MC-MRI reconstruction network, obtaining the final hard mask and reconstruction network parameters via binarization and fine-tuning. This framework demonstrates strong performance, however, its utilization of reference modality information is insufficient. These observations motivate a deeper investigation into modality-aware sampling strategies, where the under-sampling pattern is not only jointly optimized but also explicitly informed by cross-contrast dependencies.

\section{JUF-MRI Framework}\label{JUF-MRI Framework}
In this work, we present JUF-MRI, a novel framework for MC-MRI that jointly performs under-sampling optimization guided by a frequency error prior. As shown in Fig.~\ref{fig:overall_structure}, there are two key stages of the JUF-MRI framework. Firstly, the frequency error prior is obtained through a conditional diffusion model. Secondly, use this prior to jointly optimize the reconstruction network and under-sampling pattern. The specific operation for each step is introduced in the following subsections.

\subsection{Stage 1: Frequency Error Prior Generation}
In order to capture the most difficult to recover regions and the most unique features in the target modality, thereby improving the reconstruction quality, we innovatively introduce the frequency error prior. Specifically, we compute the error map between the target modality and the generated result in the frequency domain and use it as prior knowledge, which indicates which sampling points are easy or difficult to recover during MRI reconstruction. 

\begin{figure*}[htbp]
\centering
\includegraphics[width=\textwidth]{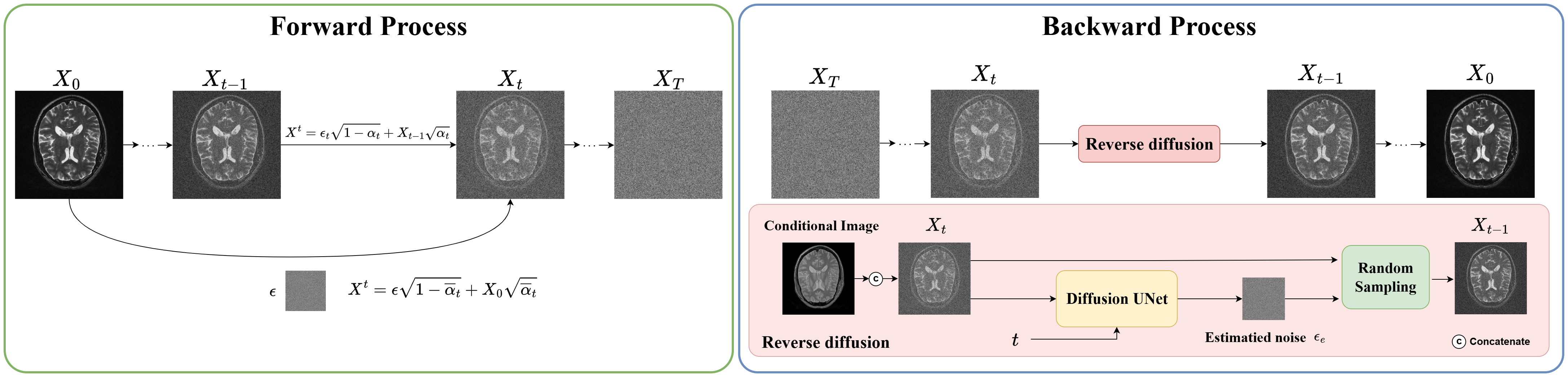}
\caption{The forward and backward processes of the proposed Conditional Diffusion Model (CDM). In the backward process, the conditional image $X_{\text{condi}}$ and the current noisy image $X_t$ are concatenated along the channel dimension. This concatenated input, along with the current timestep $t$, is then fed into the diffusion U-Net model to guide the prediction of the noise.}
\label{conditional diffusion model structure}
\end{figure*}

\subsubsection{Conditional Diffusion Model (CDM)}
We use the CDM~\cite{dhariwal2021diffusion} to generate the corresponding target modality from the reference modality, and then calculate the difference between the generated image and the real target image.

As shown in Fig.~\ref{conditional diffusion model structure}, the CDM used in this paper has a forward process and a reverse process. In the forward process, we gradually add noise to the MR images of the target modality, eventually transforming them into pure Gaussian noise. 
The formulation is as follows
\begin{equation}
 {X}_{t}={\epsilon}\sqrt{1-\overline{\alpha} _{t}}+{X}_{0}\sqrt{\overline{\alpha}_{t}} \quad {\epsilon} \sim \mathcal{N}(0, \mathbf{I}),
\label{Reparameterization}
\end{equation} 
where $X_t$ represents the image at the $t$-th time step, \( {X}_0 \) denotes the original image without noise, \( \bar{\alpha}_t = \alpha_t \alpha_{t-1} \cdots \alpha_1 \), $\alpha_t$ is used to control the intensity of noise added in the forward process.
\(  {\epsilon} \) represents Gaussian noise sampled from a standard normal distribution.

Conversely, in the reverse process, we gradually remove the noise step by step, ultimately generating a clean image. The reverse transition is modeled as a Gaussian distribution

\begin{equation}
P(X_{t-1} \mid X_t, X_0, X_{\text{condi}}) \sim \mathcal{N}(\mu, \sigma^2).
\label{reverse process}
\end{equation}

Through this process, we can synthesize the target image from $X_{T}\sim \mathcal{N}(0, \mathbf{I})$, combined with the conditional image $X_{\text{condi}}$. 
With the help of CDM, the model can generate results that closely resemble the target image. However, there is still a certain gap between the reconstructed result and the real target modality. These differences are unique to the target modality and are difficult to recover using the reference information.

\subsubsection{Frequency Error Prior (FEP)}
After synthesizing the target-modality MR image using the CDM, the FEP $r$ can be obtained by 
\begin{equation}
r = \left| \mathcal{F}\left(X_\text{sys} \right) - \mathcal{F}\left(X_\text{gt} \right) \right|,
\label{frequency error prior}
\end{equation}
where ${X_\text{sys}\in{\mathbb{R}}^{n\times n}}$ and ${X_\text{gt}\in{\mathbb{R}}^{n\times n}}$ represent the target-modality MR image synthesized by CDM and the real target-modality MR image, respectively. The variable $r\in{\mathbb{R}}^{n\times n}$  represents the absolute difference between the $k$-space representations of the generated image and the ground-truth image. 

In summary, the FEP $r$ obtained through the above operation has the following implications.
\begin{enumerate}
    \item The result of \(r\) reflects the quality of the target image synthesized by the CDM.
    
    \item A larger \(r\) indicates poorer synthesis quality and higher sampling importance, whereas a smaller \(r\) suggests easier synthesis and lower importance. Thus, \(r\) can serve as an indicator of sampling probability to a certain extent.    
\end{enumerate}
However, due to the inability of generative models to produce perfect reconstruction results, FEP \(r\) cannot fully represent the true sampling probability and needs further optimization.

\subsection{Stage 2: Joint Optimization}
After obtaining the FEP \(r\), we consider how to utilize it to optimize both the under-sampling pattern and the reconstruction network. The whole process consists of three steps, as described below:

\subsubsection{FEP-Guided Continuous Sampling Mask Acquisition}

Due to the limitations of the CDM, we follow~\cite{yang2022fast} and introduce a sampling modulation matrix $P_\text{mask}\in{\mathbb{R}}^{n\times n}$. $P_\text{mask}$ is initialized with values drawn from a uniform distribution in the range of $[-1, 1]$. Meanwhile, we use the method proposed in~\cite{bahadir2020deep,lei2024joint} to obtain the Continuous Sampling Mask (CSM)
\begin{equation}
{M}_\text{c}={\sigma}_{\beta }\left (S_{\gamma }({\sigma}_{\alpha}\left (\text{norm}(r)+{P}_\text{mask} \right ))-U\right ),
\label{continuous sampling mask}
\end{equation}
where $M_\text{c}\in{\mathbb{R}}^{n\times n}$ denotes the continuous sampling mask, ${\sigma}_{x}\left ( \cdot \right )$ is the sigmoid function with slope parameter $x$, $S_{\gamma}\left ( \cdot \right )$ represents the sparsification operation, $\gamma$ is the target sparsity level, the FEP $r$ serves as the initial sampling probability mask after normalization, and matrix $U$ is sampled from a uniform distribution over the range of $[0, 1]$. After obtaining the continuous mask $M_\text{c}$, the corresponding synthetic under-sampled image can be obtained by ${X}_{u}=\mathcal{F}^{-1}\left ({M}_\text{c}\mathcal{F}({X}_\text{gt}) \right )$, where $\mathcal{F}^{-1}$ denotes the inverse Fourier transform, and $X_\text{gt}\in{\mathbb{R}}^{n\times n}$  represents the fully-sampled target image. 

\subsubsection{Co-optimizing of CSM and Reconstruction Network}

Owing to the differentiability of the sigmoid function and the sparsification operation, the reconstruction network parameters $\theta$ and the continuous sampling mask can be jointly optimized during training. The joint optimization process can be defined as
\begin{equation}
\left \{ \hat{{P}}_\text{mask},\hat {\theta}\right \} = \underset{{P}_\text{mask},\theta}{\arg\min}\sum_{i}^{N}\left \|\varphi_{\theta}\left (X_{u}^{i}, {Y^{i}},M_\text{c}^{i}\right)-X_\text{gt}^{i} \right\|_{1} ,
\label{joint optimization proces}
\end{equation}
where $N$ is the number of training images, $\varphi_{\theta}$ denotes the reconstruction network parameterized by $\theta$, and ${Y\in{\mathbb{R}}^{n\times n}}$ represents the reference image.

\subsubsection{Binarizing CSM and Network Fine-Tuning}

After joint optimization, a fixed $P_\text{mask}$ can be obtained. Using this fixed $P_\text{mask}$ and the frequency-domain diffusion prior $r$, the continuous sampling mask for the entire training set can be derived via Eq.~\eqref{continuous sampling mask}. Since actual MRI acquisition requires a discrete sampling mask, we adopt a weighted averaging and binary search algorithm to obtain the final discrete sampling mask $M_d\in{\mathbb{R}}^{n\times n}$. The process can be formulated as 
\begin{equation}
M_d=\mathbb{I}_{\gamma }\{\frac{1}{N}\displaystyle\sum_{i}^{N}{M}_\text{c}^{i}\},
\label{discrete sampling mask}
\end{equation}
where $\mathbb{I}_{\gamma}\left ( \cdot \right )$ denotes the binary search process, whose purpose is to ensure that the discrete mask maintains a predefined mean value $\gamma$. Ultimately, the obtained discrete sampling mask $M_d$ not only incorporates information from the FEP $r$ but also participates in the joint optimization with the reconstruction network. It is worth noting that the reconstruction network parameters $\theta$ obtained during the joint optimization process are co-optimized with the continuous sampling mask $M_c$. When the continuous sampling mask $M_c$ is replaced with the discrete sampling mask $M_d$, the reconstruction network needs to be fine-tuned to adapt $\theta$ to $M_d$. The fine-tuning process can be formulated as 
\begin{equation}
\hat {\theta} = \mathop{\arg\min}\limits_{\theta}\displaystyle\sum_{i}^{N}\left \|\varphi_{\theta}\left (X_{u}^{i}, {Y^{i}},M_{d}\right)-X_\text{gt}^{i} \right\|_{1} .
\label{fine-tuned process}
\end{equation}

\subsubsection{Inference and Evaluation Pipeline}

\begin{figure}[htbp]
    \centering
    \includegraphics[width=\textwidth]{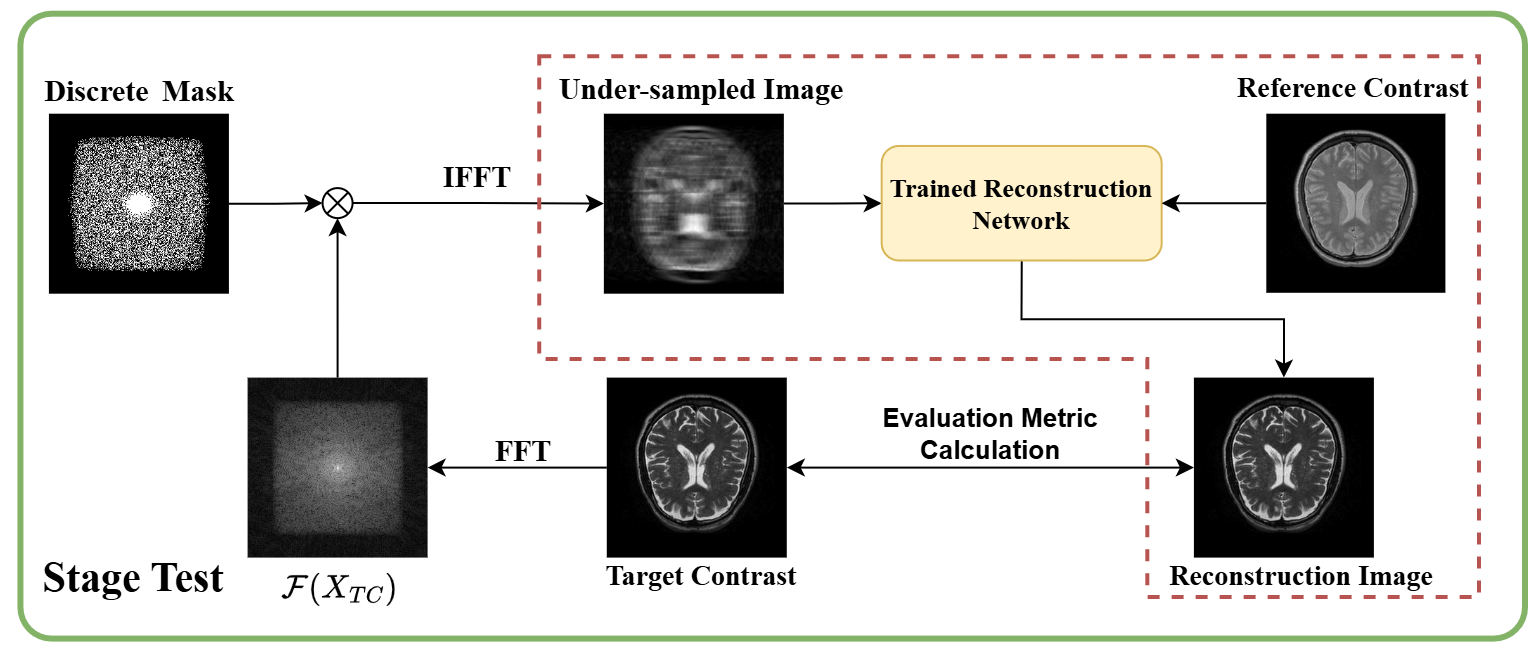}
    \caption{The testing procedure for the trained reconstruction network. The process highlighted by the red box represents the practical multi-contrast MRI reconstruction pipeline.}
    \label{fig: stage_test}
\end{figure}

Upon completing the entire training process, we obtain the final learned discrete sampling mask $M_d$ and the reconstruction network, which are co-optimized in a unified framework. The testing and evaluation pipeline is illustrated in Fig. \ref{fig: stage_test}.
To quantitatively assess our framework on the test set, we simulate the under-sampling process since prospectively acquired data with $M_d$ are unavailable. 
For each ground-truth (GT) image, we first apply FFT to obtain its $k$-space representation, which is then multiplied by the learned discrete mask $M_d$. 
The resulting masked $k$-space is transformed back to the image domain via IFFT to produce the under-sampled aliased image. This image, together with its reference contrast, is fed into the trained reconstruction network to generate the final reconstruction. Evaluation metrics are then computed between the reconstructed and original GT images.
The red box in Fig. \ref{fig: stage_test} highlights the practical inference pipeline. In a clinical setting, the MR scanner would acquire data using the optimized mask $M_d$. This prospectively under-sampled data, along with the reference contrast, would be directly input to the trained model for high-fidelity reconstruction.
It is crucial to emphasize that the conditional diffusion model is not involved during inference. Its role is limited to the training phase, where it generates the FEP that guides the joint optimization of the sampling mask and the reconstruction network. Consequently, the final inference pipeline is highly efficient and free from the computational overhead of iterative diffusion sampling.

\section{Reconstruction Network}
\subsection{Objective Function}

Classical compressed sensing algorithms reconstruct multi-contrast MR images by optimizing the energy function \eqref{equation3}. However, due to factors such as patient motion, the reference and target images in MC-MRI reconstruction are often not spatially aligned~\cite{zhang2025deep,lei2024joint}. Moreover, although the reference image contains abundant information relevant to the target image, it may also include irrelevant or uncorrelated content~\cite{lei2023decomposition}. When such information is fed into the reconstruction network, it may act as interference, preventing the network from achieving optimal performance. 

To address the above two issues, we apply a spatial transformation to the reference image to align it with the target image. Following the method in~\cite{lei2023decomposition}, we then divide the aligned reference image into two components 
\begin{equation}
Y_\text{SA}=\mathcal{T}\left ( Y,\phi \right ) = S+D,
\label{decomposition}
\end{equation}
where $\phi\in{\mathbb{R}}^{2\times n\times n}$ denotes the displacement field, $S\in{\mathbb{R}}^{ n\times n}$  and $D\in{\mathbb{R}}^{ n\times n}$ denote the components of the spatially aligned reference image that are related and unrelated to the target image, respectively. $Y_\text{SA}$ represents the spatially aligned reference image and $\mathcal{T\left ( \cdot \right )} $ represents a differentiable warping operation~\cite{jaderberg2015spatial,Xuan2022spatial,zhang2025deep}. 
Since \( S \) represents the part of the reference modality that is related to the target modality, we assume the existence of some feature transformation that can make \( S \) and \( X \) similar in some feature space, which means
\begin{equation}
AX=BS+\vartheta,
\label{feature transformation}
\end{equation}
where \( A \) and \( B \) represent two types of feature transformations, and ${\vartheta }$ represents the error. According to Eq.~\eqref{feature transformation}, we can obtain a new data fitting term that effectively avoids the issues of spatial misalignment and data noise. Moreover, the effective utilization of $k$-space data can further improve the performance of the reconstruction model~\cite{zhou2020dudornet,feng2021multi,lei2024joint}. Finally, our objective function can be organized as  
\begin{equation}
\small
\begin{aligned}
h(X,K,S,D,\phi)=&
\frac{1}{2}\left \| M\mathcal{F}(X)-\widetilde{K}\right \|_{F}^{2}
+ \frac{\gamma}{2}\left \| S + D - Y_\text{SA} \right \|_{F}^{2}
\\&
+\frac{\alpha }{2}\left \|K - \mathcal{F}(X) \right \|_{F}^{2}  + \frac{\beta }{2}\left \|AX - BS \right \|_{F}^{2}
\\&+ \lambda_{1} \mathcal{R}_{1}(X)
+ \lambda_{2} \mathcal{R}_{2}(K)
+ \lambda_{3} \psi_{1}(S) \\& + \lambda_{4} \psi_{2}(D)
+ \lambda_{5} \Phi_{1}(\phi).
\end{aligned}
\label{objective function}
\end{equation}
The first four terms are data terms. The first term is the same as in Eq.~\eqref{equation3}, the second and fourth terms are derived from Eqs.~\eqref {decomposition} and~\eqref {feature transformation}, respectively. The third term, where \( K \) represents the $k$-space data of the reconstructed target image, constrains the reconstructed $k$-space data.  
$\mathcal{R}_{1}\left ( \cdot \right )$, $\mathcal{R}_{2}\left ( \cdot \right )$, $\psi_{1}\left ( \cdot \right )$, $\psi_{2}\left ( \cdot \right )$, and $\Phi_{1}\left ( \cdot \right )$ are implicit regularization terms. $\alpha$, $\beta$, $\lambda_{i}$, $i=1, \cdots, 5$ are learnable regularization parameters that are progressively updated during the training process of the proposed reconstruction network.

\subsection{Optimization Algorithm}
We adopt the inertial block majorization minimization (TITAN) algorithm proposed in~\cite{phan2023inertial} to solve the objective function \eqref{objective function} and unroll it to construct a neural network for MC-MRI reconstruction. Compared to the iterative shrinkage-thresholding algorithm (ISTA)~\cite{daubechies2004iterative}, TITAN offers a more complete convergence guarantee for non-convex functions. 
According to the TITAN algorithm, we reformulate our objective function \eqref{objective function} to 
\begin{equation} h(X,K,S,D,\phi)=f(X,K,S,D,\phi)+\sum_{i}^{X,K,S,D,\phi}\Theta_i, 
\end{equation}
where $f$ represents the Lipschitz smooth (usually the data fitting term) function and $\Theta$ denotes the proper closed (possibly non-convex and implicit) function. Hence, in this paper, we have
\begin{equation}\label{f} 
\small
\begin{aligned}
f(X,K,S,D,\phi)=&\frac{1}{2}\left \| M\mathcal{F}(X)-\widetilde{K}\right \|_{F}^{2}
+ \frac{\gamma}{2}\left \| S + D - Y_\text{SA} \right \|_{F}^{2}\\
&+\frac{\alpha }{2}\left \|K - \mathcal{F}(X) \right \|_{F}^{2}  + \frac{\beta }{2}\left \|AX - BS \right \|_{F}^{2},
\end{aligned}
\end{equation}
where $\Theta_X(X)=\lambda_1\mathcal{R}_1(X)$, $\Theta_K(K)=\lambda_2\mathcal{R}_2(K)$, $\Theta_S(S)=\lambda_3\psi_1(S)$, $\Theta_D(D)=\lambda_4\psi_2(D)$, and 
$\Theta_\phi(\phi)=\lambda_5\Phi_1(\phi)$. Next, we will find the solution for each variable in the $t$-th iteration. 

\textbf{Update $X$:}
From the Lipschitz smooth function \eqref{f}, we have 
\begin{equation}
\small
\begin{aligned}
\nabla f(X^t)=F^*\left(F\left(X^t\right)-\tilde{K}\right)
&+\alpha^t\mathcal{F}^{-1}(\mathcal{F}\left(X^t\right)-K^t)\\
&+\beta^tA^T\left(AX^t-BS^t\right),
\end{aligned}
\label{derivative x}
\end{equation}
where $F=M\mathcal{F}$ for symbol simplicity, and $F^{*}$ is the hermitian conjunction of $F$. Hence, we update $X$ with 
\begin{equation}\label{proximal x}
X^{t+1} = \text{prox}_{\eta_x^{t}\mathcal{R}_1}(\bar{X}^t-\eta_x^{t}\nabla f(X^t)),
\end{equation}
where $\eta_x^{t}$ is the step size, the proximal operator is with 
\begin{equation}
    \text{prox}_{\gamma f}(x)=\arg\min_y g(y)+\frac{1}{2\gamma}\Vert y-x\Vert^2, 
\end{equation}
and $\bar{X^t}=X^t+\xi_x^t\left(X^t-X^{t-1}\right)$ with the inertial factor $\xi_x^t\in [0,0.5)$ for the $t$-th iteration.

\textbf{Update $\phi$:} 
According to Eqs. \eqref{decomposition} and the fact that 
\begin{equation}
    \phi^{t+1}=\text{prox}_{\eta _{\phi}\Phi_{1}}({\phi}^{t}), 
\end{equation}
we employ a spatial alignment network (SANet) to predict the spatially aligned reference contrast image $Y$ at each $t$ iterative step, which implies that 
\begin{equation}\label{SANet}
Y_\text{SA}^{t+1}=\text{SANet}(X^{t},Y)=\mathcal{T}\left( Y,\phi^{t} \right ).
\end{equation}
The detailed design of SANet will be presented in the next subsection.

\textbf{Update $K$:}
Similarly, we have 
\begin{equation}\label{derivative k}
\nabla f(K^{t})=\alpha^{t}(K^{t}-\mathcal{F}(X^{t})).
\end{equation}
Hence, we update $K$ with 
\begin{equation}\label{proximal k}
K^{t+1}=\text{prox}_{\eta_{k}^{t}\mathcal{R}_{2}}(\bar{K}^{t}-\eta_{k}^{t}\nabla f(K^{t})), 
\end{equation}
where $\bar{K}^t=K^t+\xi _k^t\left(K^t-K^{t-1}\right)$ with the inertial factor $\xi _k^t\in\left [0,0.5 \right )$ for the $t$-th iteration. 

\textbf{Update $S$:}
From \eqref{f}, we have 
\begin{equation}\label{nabS}
    \nabla f(S^{t}) = \gamma ^{t}(S^{t}+D^{t}-{Y}_\text{SA}^{t}) + \beta^{t}B^{T}(BS^{t}-AX^{t}),
\end{equation}
where $Y_\text{SA}^{t}$ is the reference image aligned with $X^{t}$ at $t$-th iteration, which can be expressed by $Y_\text{SA}^{t}=\mathcal{T}\left ( Y,\phi^{t} \right )$.
Then we update $S$ with
\begin{equation}\label{S}
    S^{t+1} = \text{prox}_{\eta_{s}^{t}\psi_{1}} ( \bar{S}^{t}-\eta_{s}^{t}\nabla f(S^{t}) ), 
\end{equation}
where $\bar{S}^t = S^t + \xi _s^t (S^t-S^{t-1} )$ with the inertial factor $\xi _s^t\in\left [0,0.5 \right )$ for the $t$-th iteration. 

\textbf{Update $D$:} Similarly, we have
\begin{equation}\label{nabD}
    \nabla f(D^{t})=\gamma^{t}(S^{t}+D^{t}-{Y}_\text{SA}^{t})),
\end{equation}
then we update $D$ with 
\begin{equation}\label{D}
    D^{t+1}=\text{prox}_{\eta _{D}\psi_{2}}\left (\bar{D}^{t}-\eta_{d}^{t}\nabla f(D^{t}) \right ), 
\end{equation}
where $\bar{D}^{t}=D^{t}+\xi_{d}^{t}(D^{t}-D^{t-1})$ with the inertial factor $\xi_d^t\in\left [0,0.5 \right )$ for the $t$-th iteration. 

According to the optimization algorithm, the latent result $X$ can be found by iteratively updating these sub-problems. In this paper, we unroll the updating to an efficient deep neural network. Next, the elaboration of the proposed scheme is given to better illustrate the overall structure. 

\subsection{Network Architecture}

\begin{figure}[htbp]
    \centering
    \includegraphics[width=\textwidth]{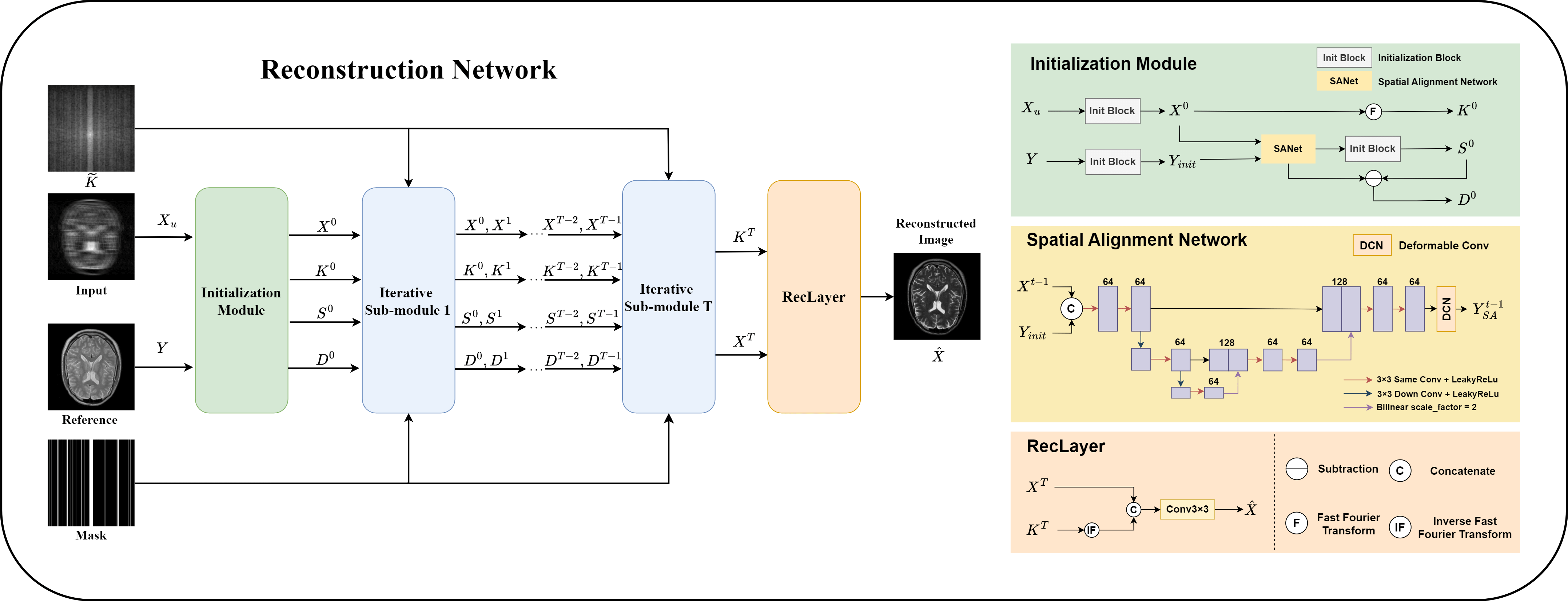}
    \caption{The architecture of the proposed deep unfolding reconstruction network, which consists of three components: the initialization module, the iterative sub-modules, and the reconstruction layer (RecLayer).}
    \label{fig:  Reconstruction network architecture}
\end{figure}
Consistent with the optimization algorithm, the architecture of the proposed reconstruction network (Fig.~\ref{fig:  Reconstruction network architecture}) consists of three components: the initialization module, the iterative sub-modules, and the reconstruction layer (RecLayer).

\subsubsection{Initialization Module}
The role of the initialization module is to preprocess parameters that need to be iterated within the network, thereby facilitating better iteration and enhancing the learning process of the model. The input of the network consists of the under-sampled target image $X_{u}$ and the reference image $Y$. Firstly, we employ the channel expansion strategy to increase the number of channels in both images to prevent information loss. Subsequently, two initialization blocks are used to map $X_{u}$ and $Y$ into the feature space, resulting in $X^{0}$ and $Y_\text{init}$. After obtaining $X^0$, we can derive $K^0$ through the FFT $K^{0}=\mathcal{F}\left ( X^{0}\right )$. We then acquire \(S^0\) and \(D^0\) by utilizing the spatial alignment network~\cite{lei2024joint} with $Y_\text{SA}^{0}=\text{SANet}(X^{0},Y_\text{init})$. After obtaining $Y_\text{SA}^{0}$, we further process it through an initialization block to derive $S^0$. Since we model the spatially aligned $Y_\text{SA}$ as composed of $S$ and $D$ in Eq.~\eqref{decomposition}, we can obtain $D^{0}=Y_\text{SA}^{0}-S^{0}$. The initialization block used for $X_u$, $Y$, and $Y_\text{SA}$ is based on ResNet~\cite{he2016deep}, consisting of ten residual blocks without batch normalization layers.

Through the processing of the initialization module, we can obtain $X^0$, $K^0$, $S^0$, $D^0$, and $Y_\text{init}$. The first four parameters 
will undergo iterative optimization in the subsequent iteration sub-module, while $Y_\text{init}$ serves as the input to the spatial alignment network within the iteration sub-module. In addition, since the TITAN algorithm requires the parameters from the previous iteration at each step, we initialize them by setting $X^{-1} = X^0$, $K^{-1} = K^0$, $S^{-1} = S^0$, $D^{-1} = D^0$. Furthermore, the parameters updated during the iterative process $\alpha$, $\beta$, $\gamma$, $\eta_x$, $\eta_k$, $\eta_s$, and $\eta_d$ are initialized to $1$, while the inertial factors $\xi_x$, $\xi_k$, $\xi_s$, and $\xi_d$ are initialized to $0.25$. This initialization ensures that all iterative sub-modules can operate consistently from the first iteration.

\subsubsection{Iterative Sub-module}

\begin{figure}[htbp]
    \centering
    \includegraphics[width=\textwidth]{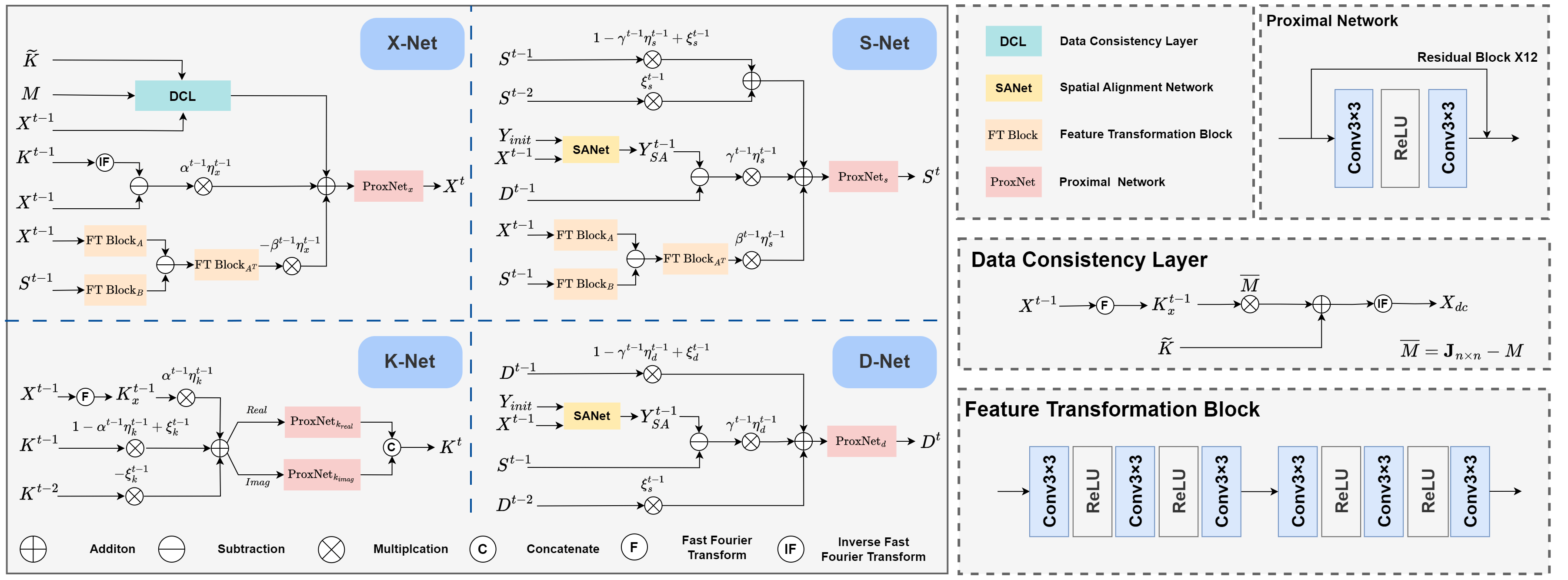}
    \caption{
    The details of the iterative sub-module in the reconstruction network. This module is responsible for the iterative optimization of four parameters: $X$, $S$, $K$, and $D$. Specifically, $K_{x}^{t-1}$ denotes the $k$-space data derived from the intermediate image $X^{t-1}$ via Fourier transform during the reconstruction process, $X_{\text{dc}}$ refers to the output of the data consistency layer applied to $X^{t-1}$, and $J_{n \times n}$ denotes an $n \times n$ matrix with every entry equal to $1$.}
    \label{fig:Iterative}
\end{figure}
As illustrated in Fig.~\ref{fig:Iterative}, the iterative sub-module of each stage consists of five components: \(X\)-Net, \(K\)-Net, \(S\)-Net, \(D\)-Net, and the Spatial Alignment Network (\(\text{SANet}\)). The first four networks are responsible for the iterative optimization of their corresponding parameters (\(X\), \(K\), \(S\), and \(D\)), while the spatial alignment network \(\text{SANet}\) plays a supporting role in the process. The following section will present a detailed description of each parameter within the iterative optimization network.

\textbf{X-Net:}
We unroll the $X$-Net based on Eqs.~\eqref{derivative x} and \eqref{proximal x}. More specifically, the linear combination of $\bar{X}^t - {{\eta_x^{t}}} F^* \left( F(X^t) - \tilde{K} \right)$ can be presented by the data consistency layer~\cite{sriram2020end,lei2024joint}; the operators $A^{T}$, $A$, and $B$ can be denoted to three different feature transformations; finally, we obtain $X^{t+1}$ from a Proximal Network (ProxNet), which was the replication of the proximal operator $\text{Prox}_{\eta_x\mathcal{R}_1}\left ( \cdot \right )$.
Note that the proximal operator, which is typically used to enforce a prior or regularization, can be replaced by a learnable deep network that implicitly captures data-dependent priors and improves performance. 
In this study, the ProxNet module is designed based on the ResNet architecture. As shown in the top-right corner of Fig. \ref{fig:Iterative}, ProxNet consists of 12 lightweight residual blocks, each containing two $3\times 3$ convolutional layers.
A ReLU activation follows the first convolution, and a residual connection links the block input and output. All convolutional layers are initialized using Kaiming Normal initialization ($a =0$, mode=`fan\_in'), and the biases are set to zero. The entire ProxNet module contains only $0.222$ M trainable parameters, which contributes substantially to the overall parameter efficiency of the proposed deep unfolding network.

\textbf{K-Net:} 
Similar to \( X \), we unroll the $K$-Net based on Eqs. \eqref{derivative k} and \eqref{proximal k}. 
Since \( K \) is complex-valued, it consists of two channels: one for the real part and one for the imaginary part. We use two distinct ProxNets to predict the proximal operators for each part separately.

\textbf{S-Net \& D-Net:}
The construction processes of $S$-Net and $D$-Net are similar to those of $X$-Net and $K$-Net, respectively. We unroll the $S$-Net based on Eqs. \eqref{nabS} and \eqref{S} and $D$-Net based on Eqs. \eqref{nabD} and \eqref{D}.

\textbf{SANet:} We employ the Deformable Adaptive Sampler (DAS) module~\cite{lei2024joint} as the spatial alignment network. The DAS module leverages deformable convolutions to estimate multiple spatial offsets at each location, enabling more flexible and precise alignment compared to traditional one-to-one spatial resampling approaches. 

\subsubsection{The Reconstruction Layer (RecLayer)}
After \( T \) iterations, we obtain the reconstructed image \( X^T \) and the $k$-space data \( K^T \). A reconstruction layer is then used to balance the contributions from the image domain and the frequency domain, and reduce the number of channels to one in order to generate the final image. Hence, the final reconstructed image $\hat{X}$ can be obtained by
\begin{equation}
\hat{X}= \text{RecLayer}(X^{T},\mathcal{F}^{-1}(K^{T}))
\label{wal}. 
\end{equation}

\subsubsection{Loss Function}
A critical limitation of existing MRI reconstruction approaches is their insufficient attention to high-frequency components, which substantially degrades fine structural recovery~\cite{fang2025digging}. To address this issue, we propose a combined frequency-domain loss for our proposed deep unfolding network, replacing the traditional $\ell_1$ and $\ell_2$ losses to encourage the network to better learn high-frequency features. The combined frequency-domain loss can be formulated as
\begin{equation}
\left\{
\begin{aligned}
&X_\text{rec\_{low}}=\text{Guass\_kernel}\left ( \hat{X}\right ), \\
&X_\text{rec\_{high}}=\left|\hat{X}-\text{Guass\_kernel}\left ( \hat{X}\right )\right|, \\
&X_\text{gt\_{low}}=\text{Guass\_kernel}\left ( {X}_\text{gt}\right ), \\
&X_\text{gt\_{high}}=\left|X_\text{gt}-\text{Guass\_kernel}\left ( {X}_\text{gt}\right )\right|, \\
&\ell_\text{fre}= \lambda_{1} \left| X_\text{rec\_{low}} - X_\text{gt\_{low}} \right|+ \lambda_{2} \left| X_\text{rec\_{high}} - X_\text{gt\_{high}} \right|, \\
& \ell_1= \Vert\hat{X}-X_\text{gt}\Vert_{1}, \\
&\ell_\text{final} = \alpha \ell_\text{fre} + \beta \ell_{1},
\end{aligned}
\right.\nonumber
\label{combined frequency-domain loss}
\end{equation}
where $ \text{Gauss\_kernel} (\cdot)$ denotes the convolution operation with a Gaussian blur kernel. 
The output $X_\text{rec\_{low}}$ of $ \text{Gauss\_kernel} (\cdot)$  represents the low-frequency component of reconstructed image $\hat{X}$. After obtaining $X_\text{rec\_{low}}$, the high-frequency component $X_\text{rec\_{high}}$ is obtained by subtracting the $X_\text{rec\_{low}}$ from the $\hat{X}$. The same strategy applies to the ground truth image ${X}_\text{gt}$ as well. Afterward, we obtain the frequency loss $\ell_\text{fre}$ by combining the subtraction of the low-frequency and high-frequency components of $\hat{X}$ and ${X}_\text{gt}$ with the trade-off coefficients $\lambda_1$ and $\lambda_2$. The final loss $\ell_\text{final}$ is composed of $\ell_\text{fre}$ and $\ell_1$, where $\alpha$ and $\beta$ are the weights for these two losses, respectively. 

\section{Experiments} 
To validate the efficacy of the proposed JUF-MRI framework, we conduct comprehensive evaluations on three MC-MRI datasets that cover different modalities. Our experiments are structured into two components: (1) an isolated evaluation of the performance of the reconstruction network, and (2) an assessment of the FEP-guided joint optimization framework. 

\subsection{Datasets}
We trained and evaluated our proposed method using three datasets, including IXI\footnote{\url{http://brain-development.org/ixi-dataset/}}, BraTS2018\footnote{\url{https://www.med.upenn.edu/sbia/brats2018/data.html}}, and FastMRI\footnote{\url{https://fastmri.med.nyu.edu/ }}. 
Following standard protocols in the field, we processed each volumetric dataset by extracting the central 20 slices to form our final experimental data. The datasets were partitioned into training, validation, and test subsets using a $7:1:2$ ratio. Specifically, we utilized 570 volumes from the IXI dataset, where the PD-weighted modality (resolution $256 \times 256$) served as reference for T2-weighted modality reconstruction; the Brats2018 Dataset contributed 285 volumes with T1-weighted modality guiding T2-weighted reconstruction (resolution $240 \times 240$); and the FastMRI Dataset provided 272 volumes where PD-weighted modality guided FSPD-weighted reconstruction, with images resized from original $320 \times 320$ to $256 \times 256$ resolution.

\subsection{Implementation Details}
Our experiments were implemented using PyTorch on NVIDIA GeForce RTX 3090 GPUs. The conditional diffusion model was trained with an Adam optimizer (fixed learning rate of \(2 \times 10^{-5}\), no weight decay) incorporating exponential moving average (weight=$0.999$) for training stability, running for $300$ epochs with \( T = 1000 \) diffusion steps. For the reconstruction model, we employed Adam optimization (learning rate \(10^{-4}\)) with batch size $2$ across $30$ training epochs. We evaluated two under-sampling mask types (1D equispaced and learned) at four acceleration rates (4$\times$, 8$\times$, 10$\times$, 30$\times$) with corresponding central sampling fractions ($0.084$, $0.042$, $0.032$, $0.0125$), where higher acceleration rates enable faster acquisition but intensify reconstruction challenges due to increased $k$-space sparsity. Quantitative evaluation was performed using three standard metrics: PSNR, SSIM, and RMSE. The standard deviation of each metric was also tested.

\begin{sidewaystable*}[p]
\centering

\renewcommand{\arraystretch}{1.75} 

\setlength{\tabcolsep}{1.2mm} 

\caption{Quantitative results of reconstruction T2-weighted modality with PD-weighted reference on the IXI dataset under different acceleration rates. `Equispaced' and `Learned' indicate the mask generation methods.}
\label{tab:ixi_results}

\resizebox{\textwidth}{!}{
    \begin{tabular}{c|c|cccccccccccc} 
    \toprule
    \multirow{2}{*}{\textbf{IXI}} & \multirow{2}{*}{\textbf{Methods}} &
    \multicolumn{3}{c}{4$\times$Acceleration} & \multicolumn{3}{c}{8$\times$Acceleration} &
    \multicolumn{3}{c}{10$\times$Acceleration} & \multicolumn{3}{c}{30$\times$Acceleration} \\
    \cmidrule(lr){3-5} \cmidrule(lr){6-8} \cmidrule(lr){9-11} \cmidrule(lr){12-14}
    & & PSNR $\uparrow$ & SSIM $\uparrow$ & RMSE $\downarrow$ &
    PSNR $\uparrow$ & SSIM $\uparrow$ & RMSE $\downarrow$ &
    PSNR $\uparrow$ & SSIM $\uparrow$ & RMSE $\downarrow$ &
    PSNR $\uparrow$ & SSIM $\uparrow$ & RMSE $\downarrow$ \\
    \midrule
    \multirow{9}{*}{\rotatebox{90}{Equispaced}} 
    & Zero-Filled (ZF) & 25.98\,{$\pm$}\,2.00 & 0.5918\,{$\pm$}\,0.0445 & 13.13\,{$\pm$}\,2.70 & 23.78\,{$\pm$}\,2.02 & 0.5124\,{$\pm$}\,0.0526 & 16.93\,{$\pm$}\,3.54 & 23.29\,{$\pm$}\,2.02 & 0.5134\,{$\pm$}\,0.0550 & 17.89\,{$\pm$}\,3.76 & 22.08\,{$\pm$}\,2.03 & 0.4435\,{$\pm$}\,0.0565 & 20.57\,{$\pm$}\,4.33 \\
    & UNet\cite{ronneberger2015u} & 33.04\,{$\pm$}\,2.23 & 0.9173\,{$\pm$}\,0.0222 &5.86\,{$\pm$}\,1.42 & 28.85\,{$\pm$}\,2.18 & 0.8531\,{$\pm$}\,0.0335 & 9.48\,{$\pm$}\,2.27 & 27.38\,{$\pm$}\,2.14 & 0.8255\,{$\pm$}\,0.0366 & 11.23\,{$\pm$}\,2.67 & 23.73\,{$\pm$}\,2.02 & 0.7141\,{$\pm$}\,0.0449 & 17.04\,{$\pm$}\,3.76 \\
    & Multi-UNet & 37.74\,{$\pm$}\,2.20 & 0.9647\,{$\pm$}\,0.0121 & 3.41\,{$\pm$}\,0.80 & 35.60\,{$\pm$}\,2.19 & 0.9566\,{$\pm$}\,0.0149& 4.36\,{$\pm$}\,1.03 &35.33\,{$\pm$}\,2.22& 0.9562\,{$\pm$}\,0.0149 & 4.50\,{$\pm$}\,1.08 & 34.30\,{$\pm$}\,2.23 & 0.9521\,{$\pm$}\,0.0162 & 5.07\,{$\pm$}\,1.25 \\
    & CUNet\cite{deng2020deep} &36.93\,{$\pm$}\,1.90& 0.9596\,{$\pm$}\,0.0124 &3.72\,{$\pm$}\,0.78  &35.32\,{$\pm$}\,1.97  &0.9507\,{$\pm$}\,0.0148  &4.48\,{$\pm$}\,0.98  & 34.94\,{$\pm$}\,2.02 & 0.9498\,{$\pm$}\,0.0157 &4.69\,{$\pm$}\,1.06  &33.86\,{$\pm$}\,2.04 & 0.9398\,{$\pm$}\,0.0170 &  5.31\,{$\pm$}\,1.23   \\
    & MDUNet\cite{xiang2018deep}  & 38.54\,{$\pm$}\,2.39 & 0.9693\,{$\pm$}\,0.0113 & 3.13\,{$\pm$}\,0.82  & 37.76\,{$\pm$}\,2.32 & 0.9657\,{$\pm$}\,0.0128  &3.42\,{$\pm$}\,0.86  &37.39\,{$\pm$}\,2.33  &0.9650\,{$\pm$}\,0.0135  &3.56\,{$\pm$}\,0.91  &36.66\,{$\pm$}\,2.29  &0.9611\,{$\pm$}\,0.0135  &3.87\,{$\pm$}\,0.98  \\
    & Restormer\cite{zamir2022restormer} & 40.88\,{$\pm$}\,2.60 & 0.9777\,{$\pm$}\,0.0083 & 2.40\,{$\pm$}\,0.65  & 38.83\,{$\pm$}\,2.38 & 0.9715\,{$\pm$}\,0.0112  &3.04\,{$\pm$}\,0.82  &38.42\,{$\pm$}\,2.51  &0.9708\,{$\pm$}\,0.0114  &3.18\,{$\pm$}\,0.85  &37.41\,{$\pm$}\,2.50  &0.9677\,{$\pm$}\,0.0131  &3.57\,{$\pm$}\,0.96  \\
    & VANet\cite{lei2023decomposition} & 42.18\,{$\pm$}\,2.40 & 0.9811\,{$\pm$}\,0.0067 & 2.06\,{$\pm$}\,0.52 & 39.77\,{$\pm$}\,2.50 & 0.9741\,{$\pm$}\,0.0100 & 2.72\,{$\pm$}\,0.73 & 38.16\,{$\pm$}\,2.44 & 0.9657\,{$\pm$}\,0.0119 & 3.27\,{$\pm$}\,0.86 & 37.63\,{$\pm$}\,2.52 & 0.9649\,{$\pm$}\,0.0126 & 3.49\,{$\pm$}\,0.96\\
    & MC-DuDoN\cite{lei2024joint} & \underline{45.58\,{$\pm$}\,2.71} & \underline{0.9900\,{$\pm$}\,0.0040} & \underline{1.40\,{$\pm$}\,0.40} & \underline{40.94\,{$\pm$}\,2.56} & \underline{0.9768\,{$\pm$}\,0.0086} & \underline{2.38\,{$\pm$}\,0.66} & \underline{40.09\,{$\pm$}\,2.60} & \underline{0.9757\,{$\pm$}\,0.0094} & \underline{2.63\,{$\pm$}\,0.74} & \underline{37.74\,{$\pm$}\,2.50} & \underline{0.9676\,{$\pm$}\,0.0130} & \underline{3.44\,{$\pm$}\,0.94} \\
    & \textbf{JUF-MRI (Ours)} & \textbf{45.89\,{$\pm$}\,2.71} & \textbf{0.9905\,{$\pm$}\,0.0037} & \textbf{1.36\,{$\pm$}\,0.39} & \textbf{41.55\,{$\pm$}\,2.64} & \textbf{0.9782\,{$\pm$}\,0.0079} & \textbf{2.23\,{$\pm$}\,0.63} & \textbf{40.47\,{$\pm$}\,2.63} & \textbf{0.9778\,{$\pm$}\,0.0091} & \textbf{2.38\,{$\pm$}\,0.72} & \textbf{37.86\,{$\pm$}\,2.47} & \textbf{0.9693\,{$\pm$}\,0.0125}  
    & \textbf{3.39\,{$\pm$}\,0.91} \\
    \midrule
    \multirow{3}{*}{\rotatebox{90}{Learned}}
    & MC-DuDoN (LOUPE) & 52.62\,{$\pm$}\,1.79& 0.9960\,{$\pm$}\,0.0012 & 0.61\,{$\pm$}\,0.12 & 48.36\,{$\pm$}\,2.34 & 0.9925\,{$\pm$}\,0.0024 & 1.01\,{$\pm$}\,0.26 & \underline{47.29\,{$\pm$}\,2.57} & \underline{0.9912\,{$\pm$}\,0.0030} & \underline{1.15\,{$\pm$}\,0.31} & 39.51\,{$\pm$}\,2.19 & 0.9721\,{$\pm$}\,0.0103 & 2.78\,{$\pm$}\,0.66 \\
    & JUF-MRI (LOUPE) & \underline{53.91\,{$\pm$}\,1.80} & \underline{0.9962\,{$\pm$}\,0.0013} & \underline{0.53\,{$\pm$}\,0.12} & \underline{48.75\,{$\pm$}\,2.45} & \underline{0.9930\,{$\pm$}\,0.0025} & \underline{0.96\,{$\pm$}\,0.26} & 47.14\,{$\pm$}\,2.65 & 0.9905\,{$\pm$}\,0.0032& 1.17\,{$\pm$}\,0.34 & \underline{40.62\,{$\pm$}\,2.45} & \underline{0.9745\,{$\pm$}\,0.0089} & \underline{2.47\,{$\pm$}\,0.63}\\
    & \textbf{JUF-MRI (Ours)} & \textbf{54.23\,{$\pm$}\,1.82} & \textbf{0.9966\,{$\pm$}\,0.0011} & \textbf{0.51\,{$\pm$}\,0.10} & \textbf{49.35\,{$\pm$}\,2.41} & \textbf{0.9937\,{$\pm$}\,0.0022} & \textbf{0.90\,{$\pm$}\,0.24} & \textbf{47.46\,{$\pm$}\,2.62} & \textbf{0.9916\,{$\pm$}\,0.0030} & \textbf{1.13\,{$\pm$}\,0.32} & \textbf{41.62\,{$\pm$}\,2.62} & \textbf{0.9758\,{$\pm$}\,0.0075} & \textbf{2.21\,{$\pm$}\,0.60} \\
    \bottomrule
    \end{tabular}
}
\end{sidewaystable*}

\begin{sidewaystable*}[p]
\centering

\renewcommand{\arraystretch}{1.75} 

\setlength{\tabcolsep}{1.2mm} 

\caption{Quantitative results of reconstruction T2-weighted modality with T1-weighted reference on the Brats2018 dataset under different acceleration rates. `Equispaced' and `Learned' indicate the mask generation methods.}
\label{tab:brats_results}

\resizebox{\textwidth}{!}{
    \begin{tabular}{c|c|cccccccccccc} 
    \toprule
    \multirow{2}{*}{\textbf{Brats2018}} & \multirow{2}{*}{\textbf{Methods}} & \multicolumn{3}{c}{4$\times$Acceleration} & \multicolumn{3}{c}{8$\times$Acceleration} & \multicolumn{3}{c}{10$\times$Acceleration} & \multicolumn{3}{c}{30$\times$Acceleration} \\
    \cmidrule(lr){3-5} \cmidrule(lr){6-8} \cmidrule(lr){9-11} \cmidrule(lr){12-14}
    & & PSNR $\uparrow$ & SSIM $\uparrow$ & RMSE $\downarrow$ & PSNR $\uparrow$ & SSIM $\uparrow$ & RMSE $\downarrow$ & PSNR $\uparrow$ & SSIM $\uparrow$ & RMSE $\downarrow$ & PSNR $\uparrow$ & SSIM $\uparrow$ & RMSE $\downarrow$ \\
    \midrule
    \multirow{9}{*}{\rotatebox{90}{Equispaced}} 
    & Zero-Filled (ZF) & 28.34\,{$\pm$}\,1.70 & 0.6080\,{$\pm$}\,0.0418 & 9.93\,{$\pm$}\,1.78 & 24.87\,{$\pm$}\,1.62 & 0.5308\,{$\pm$}\,0.0430 & 14.79\,{$\pm$}\,2.49 & 24.05\,{$\pm$}\,1.61 & 0.5139\,{$\pm$}\,0.0423 & 16.25\,{$\pm$}\,2.75 & 22.17\,{$\pm$}\,1.52 & 0.4634\,{$\pm$}\,0.0396 & 20.15\,{$\pm$}\,3.32 \\
    & UNet\cite{ronneberger2015u} & 34.97\,{$\pm$}\,1.70 & 0.9506\,{$\pm$}\,0.0118 & 4.64\,{$\pm$}\,0.89 & 30.27\,{$\pm$}\,1.74 & 0.9020\,{$\pm$}\,0.0177 & 7.98\,{$\pm$}\,1.55 & 28.93\,{$\pm$}\,1.79 & 0.8862\,{$\pm$}\,0.0209 & 9.32\,{$\pm$}\,1.88 & 25.03\,{$\pm$}\,1.77 & 0.8129\,{$\pm$}\,0.0283 & 14.58\,{$\pm$}\,2.87 \\
    & Multi-UNet & 36.62\,{$\pm$}\,1.75 & 0.9661\,{$\pm$}\,0.0086 & 3.84\,{$\pm$}\,0.74 & 32.85\,{$\pm$}\,1.62 & 0.9444\,{$\pm$}\,0.0144 & 5.91\,{$\pm$}\,1.08 & 32.06\,{$\pm$}\,1.64 & 0.9401\,{$\pm$}\,0.0157 & 6.48\,{$\pm$}\,1.21 & 30.11\,{$\pm$}\,1.80 & 0.9237\,{$\pm$}\,0.0190 & 8.13\,{$\pm$}\,1.72 \\
    & CUNet\cite{deng2020deep} & 36.63\,{$\pm$}\,1.77 & 0.9648\,{$\pm$}\,0.0187 & 3.84\,{$\pm$}\,0.78 & 33.41\,{$\pm$}\,1.63 & 0.9470\,{$\pm$}\,0.0199 & 5.54\,{$\pm$}\,1.02 & 32.75\,{$\pm$}\,1.61 & 0.9425\,{$\pm$}\,0.0221 & 5.98\,{$\pm$}\,1.10 & 30.15\,{$\pm$}\,1.65 & 0.9244\,{$\pm$}\,0.0206 & 8.07\,{$\pm$}\,1.57 \\
    & MDUNet\cite{xiang2018deep} & 37.50\,{$\pm$}\,1.72 & 0.9709\,{$\pm$}\,0.0079 & 3.47\,{$\pm$}\,0.67 & 34.59\,{$\pm$}\,1.75 & 0.9565\,{$\pm$}\,0.0127 & 4.85\,{$\pm$}\,0.95 & 33.95\,{$\pm$}\,1.79 & 0.9526\,{$\pm$}\,0.014 & 5.22\,{$\pm$}\,1.05 & 31.66\,{$\pm$}\,1.91 & 0.9389\,{$\pm$}\,0.0187 & 6.83\,{$\pm$}\,1.53 \\
    & Restormer\cite{zamir2022restormer} & 38.72\,{$\pm$}\,2.11 & 0.9750\,{$\pm$}\,0.0074 & 3.04\,{$\pm$}\,0.71 & 35.22\,{$\pm$}\,2.04 & 0.9584\,{$\pm$}\,0.0130 & 4.54\,{$\pm$}\,1.04 & 34.38\,{$\pm$}\,2.01 & 0.9547\,{$\pm$}\,0.0143 & 5.00\,{$\pm$}\,1.14 & 31.81\,{$\pm$}\,2.04 & 0.9385\,{$\pm$}\,0.0195 & 6.73\,{$\pm$}\,1.68 \\
    & VANet\cite{lei2023decomposition} & 41.01\,{$\pm$}\,1.95 & 0.9842\,{$\pm$}\,0.0049 & 2.33\,{$\pm$}\,0.51 & 36.55\,{$\pm$}\,1.88 & 0.9673\,{$\pm$}\,0.0109 & 3.88\,{$\pm$}\,0.82& 35.00\,{$\pm$}\,1.87 & 0.9585\,{$\pm$}\,0.0133 & 4.64\,{$\pm$}\,0.98 & 32.02\,{$\pm$}\,1.90 & 0.9399\,{$\pm$}\,0.0182 & 6.54\,{$\pm$}\,1.46 \\
    & MC-DuDoN\cite{lei2024joint} & \underline{44.79\,{$\pm$}\,2.19} & \underline{0.9927\,{$\pm$}\,0.0025} & \underline{1.52\,{$\pm$}\,0.38} & \textbf{39.29\,{$\pm$}\,2.08} & \textbf{0.9796\,{$\pm$}\,0.0077} & \textbf{2.85\,{$\pm$}\,0.67} & \textbf{38.29\,{$\pm$}\,2.14} & \textbf{0.9752\,{$\pm$}\,0.0095} & \textbf{3.20\,{$\pm$}\,0.78} & \underline{33.07\,{$\pm$}\,2.11} & \underline{0.9481\,{$\pm$}\,0.0181} & \underline{5.84\,{$\pm$}\,1.49} \\
    & \textbf{JUF-MRI (Ours)} & \textbf{46.05\,{$\pm$}\,2.34} & \textbf{0.9945\,{$\pm$}\,0.0021} & \textbf{1.32\,{$\pm$}\,0.35} & \underline{39.12\,{$\pm$}\,2.10} & \underline{0.9785\,{$\pm$}\,0.0085} & \underline{2.90\,{$\pm$}\,0.69} & \underline{37.83\,{$\pm$}\,2.12} & \underline{0.9737\,{$\pm$}\,0.0105} & \underline{3.36\,{$\pm$}\,0.81} & \textbf{33.15\,{$\pm$}\,2.13} & \textbf{0.9493\,{$\pm$}\,0.0189} & \textbf{5.78\,{$\pm$}\,1.50} \\

    \midrule
    \multirow{3}{*}{\rotatebox{90}{Learned}}
    & MC-DuDoN (LOUPE) & 51.06\,{$\pm$}\,2.19 & 0.9980\,{$\pm$}\,0.0026& 0.74\,{$\pm$}\,0.24 & 46.61\,{$\pm$}\,2.28 & 0.9944\,{$\pm$}\,0.0030 & 1.23\,{$\pm$}\,0.33 & 44.99\,{$\pm$}\,2.20 & 0.9928\,{$\pm$}\,0.0036 & 1.48\,{$\pm$}\,0.39 & 36.60\,{$\pm$}\,1.89 & 0.9678\,{$\pm$}\,0.0100 & 3.86\,{$\pm$}\,0.83 \\
    & JUF-MRI (LOUPE) & \underline{56.89\,{$\pm$}\,2.13} & \underline{0.9995\,{$\pm$}\,0.0003} & \underline{0.38\,{$\pm$}\,0.09} & \underline{48.40\,{$\pm$}\,2.58} & \underline{0.9961\,{$\pm$}\,0.0023} & \underline{1.01\,{$\pm$}\,0.31} & \underline{46.03\,{$\pm$}\,2.49} & \underline{0.9940\,{$\pm$}\,0.0035} & \underline{1.33\,{$\pm$}\,0.40} & \underline{39.80\,{$\pm$}\,2.12} & \underline{0.9816\,{$\pm$}\,0.0064} & \underline{2.69\,{$\pm$}\,0.64} \\
    & \textbf{JUF-MRI (Ours)} & \textbf{58.46\,{$\pm$}\,1.83} & \textbf{0.9996\,{$\pm$}\,0.0002} & \textbf{0.31\,{$\pm$}\,0.07} & \textbf{48.49\,{$\pm$}\,2.76} & \textbf{0.9966\,{$\pm$}\,0.0018} & \textbf{1.01\,{$\pm$}\,0.33} & \textbf{46.42\,{$\pm$}\,2.75} & \textbf{0.9951\,{$\pm$}\,0.0025} & \textbf{1.28\,{$\pm$}\,0.40} & \textbf{40.39\,{$\pm$}\,2.33} & \textbf{0.9849\,{$\pm$}\,0.0058} & \textbf{2.52\,{$\pm$}\,0.66} \\
    \bottomrule
    \end{tabular}
}
\end{sidewaystable*}

\begin{sidewaystable*}[p]
\centering

\renewcommand{\arraystretch}{1.75}

\setlength{\tabcolsep}{1.2mm} 

\caption{Quantitative results of reconstruction T2-weighted modality with PD-weighted reference on the FastMRI dataset under different acceleration rates. `Equispaced' and `Learned' indicate the mask generation methods.}
\label{tab:fastmri_results}

\resizebox{\textwidth}{!}{
    \begin{tabular}{c|c|cccccccccccc} 
    \toprule
    \multirow{2}{*}{\textbf{FastMRI}} & \multirow{2}{*}{\textbf{Methods}} & \multicolumn{3}{c}{4$\times$Acceleration} & \multicolumn{3}{c}{8$\times$Acceleration} & \multicolumn{3}{c}{10$\times$Acceleration} & \multicolumn{3}{c}{30$\times$Acceleration} \\
    \cmidrule(lr){3-5} \cmidrule(lr){6-8} \cmidrule(lr){9-11} \cmidrule(lr){12-14}
    & & PSNR $\uparrow$ & SSIM $\uparrow$ & RMSE $\downarrow$ & PSNR $\uparrow$ & SSIM $\uparrow$ & RMSE $\downarrow$ & PSNR $\uparrow$ & SSIM $\uparrow$ & RMSE $\downarrow$ & PSNR $\uparrow$ & SSIM $\uparrow$ & RMSE $\downarrow$ \\
    \midrule
    \multirow{9}{*}{\rotatebox{90}{Equispaced}} 
    & Zero-Filled (ZF) & 23.96\,{$\pm$}\,2.20 & 0.5753\,{$\pm$}\,0.0583 & 16.66\,{$\pm$}\,3.91 & 21.17\,{$\pm$}\,1.98 & 0.4183\,{$\pm$}\,0.0747 & 22.84\,{$\pm$}\,4.92 & 20.44\,{$\pm$}\,1.93 & 0.3763\,{$\pm$}\,0.0760 & 24.82\,{$\pm$}\,5.23 & 18.12\,{$\pm$}\,1.85 & 0.2761\,{$\pm$}\,0.0850 & 32.36\,{$\pm$}\,6.55 \\
    & UNet\cite{ronneberger2015u} & 26.47\,{$\pm$}\,2.26 & 0.6939\,{$\pm$}\,0.0401 & 12.51\,{$\pm$}\,3.15 & 23.32\,{$\pm$}\,2.06 & 0.5186\,{$\pm$}\,0.0618 & 17.89\,{$\pm$}\,4.22 & 21.82\,{$\pm$}\,1.88 & 0.4560\,{$\pm$}\,0.0669 & 21.16\,{$\pm$}\,4.56 & 19.44\,{$\pm$}\,1.85 & 0.3163\,{$\pm$}\,0.0871 & 27.81\,{$\pm$}\,5.77 \\
    & Multi-UNet & 26.85\,{$\pm$}\,2.16 & 0.7095\,{$\pm$}\,0.0373 & 11.94\,{$\pm$}\,2.89 & 24.37\,{$\pm$}\,1.86 & 0.5564\,{$\pm$}\,0.0559 & 15.78\,{$\pm$}\,3.42 & 23.49\,{$\pm$}\,1.72 & 0.5081\,{$\pm$}\,0.0613 & 17.41\,{$\pm$}\,3.52 & 21.76\,{$\pm$}\,1.63 & 0.4083\,{$\pm$}\,0.0761 & 21.20\,{$\pm$}\,4.08 \\
    & CUNet\cite{deng2020deep} & 26.86\,{$\pm$}\,2.16 & 0.7103\,{$\pm$}\,0.0361 & 11.93\,{$\pm$}\,2.88 & 24.20\,{$\pm$}\,1.91 & 0.5494\,{$\pm$}\,0.0549 & 16.11\,{$\pm$}\,3.56 & 23.49\,{$\pm$}\,1.76 & 0.5093\,{$\pm$}\,0.0598 & 17.42\,{$\pm$}\,3.58 & 21.32\,{$\pm$}\,1.61 & 0.3881\,{$\pm$}\,0.0744 & 22.29\,{$\pm$}\,4.22 \\
    & MDUNet\cite{xiang2018deep} & 27.04\,{$\pm$}\,2.18 & 0.7088\,{$\pm$}\,0.0379 & 11.69\,{$\pm$}\,2.84 & 25.10\,{$\pm$}\,1.96 & 0.5714\,{$\pm$}\,0.0586 & 14.54\,{$\pm$}\,3.27 & 24.48\,{$\pm$}\,1.89 & 0.5326\,{$\pm$}\,0.0642 & 15.59\,{$\pm$}\,3.42 & 22.85\,{$\pm$}\,1.74 & 0.4193\,{$\pm$}\,0.0805 & 18.75\,{$\pm$}\,3.82 \\ 
    & Restormer\cite{zamir2022restormer} & 27.52\,{$\pm$}\,2.24 & 0.7272\,{$\pm$}\,0.0370 & 11.08\,{$\pm$}\,2.75 & 25.47\,{$\pm$}\,2.03 & 0.5916\,{$\pm$}\,0.0578 & 13.95\,{$\pm$}\,3.23 & 24.88\,{$\pm$}\,1.94 & 0.5542\,{$\pm$}\,0.0629 & 14.91\,{$\pm$}\,3.33 & \underline{23.52\,{$\pm$}\,1.77} & \underline{0.4603\,{$\pm$}\,0.0781} & \underline{17.35\,{$\pm$}\,3.58} \\
    & VANet\cite{lei2023decomposition} & 27.94\,{$\pm$}\,2.25 & 0.7464\,{$\pm$}\,0.0361 & 10.56\,{$\pm$}\,2.63 & 25.62\,{$\pm$}\,2.01 & 0.6009\,{$\pm$}\,0.0565 & 13.71\,{$\pm$}\,3.15 & 24.64\,{$\pm$}\,1.90 & 0.5472\,{$\pm$}\,0.0628 & 15.31\,{$\pm$}\,3.37 & 23.11\,{$\pm$}\,1.77 & 0.4336\,{$\pm$}\,0.0779 & 18.21\,{$\pm$}\,3.78 \\
    & MC-DuDoN\cite{lei2024joint} & \underline{28.15\,{$\pm$}\,2.26} & \underline{0.7540\,{$\pm$}\,0.0366} & \underline{10.30\,{$\pm$}\,2.58} & \underline{25.89\,{$\pm$}\,2.06} & \underline{0.6114\,{$\pm$}\,0.0575} & \underline{13.30\,{$\pm$}\,3.11} & \underline{25.07\,{$\pm$}\,1.94} & \underline{0.5659\,{$\pm$}\,0.0630} & \underline{14.58\,{$\pm$}\,3.26} & 23.33\,{$\pm$}\,1.77 & 0.4514\,{$\pm$}\,0.0786 & 17.74\,{$\pm$}\,3.68 \\
    & \textbf{JUF-MRI (Ours)} & \textbf{28.19\,{$\pm$}\,2.26} & \textbf{0.7557\,{$\pm$}\,0.0362} & \textbf{10.26\,{$\pm$}\,2.57} & \textbf{25.94\,{$\pm$}\,2.06} & \textbf{0.6125\,{$\pm$}\,0.0573} & \textbf{13.22\,{$\pm$}\,3.09} & \textbf{25.17\,{$\pm$}\,1.94} & \textbf{0.5672\,{$\pm$}\,0.0634} & \textbf{14.42\,{$\pm$}\,3.23} & \textbf{23.67\,{$\pm$}\,1.79} & \textbf{0.4606\,{$\pm$}\,0.0785} & \textbf{17.08\,{$\pm$}\,3.57} \\
    \midrule
    \multirow{3}{*}{\rotatebox{90}{Learned}}
    & MC-DuDoN (LOUPE) & 29.27\,{$\pm$}\,2.30 & 0.8042\,{$\pm$}\,0.0289 & 9.07\,{$\pm$}\,2.31 & 27.28\,{$\pm$}\,2.26 & 0.6731\,{$\pm$}\,0.0545 & 11.39\,{$\pm$}\,2.84 & 26.87\,{$\pm$}\,2.26 & 0.6388\,{$\pm$}\,0.0610 & 11.94\,{$\pm$}\,2.97 & \underline{24.38\,{$\pm$}\,1.97} & \underline{0.4962\,{$\pm$}\,0.0771} & \underline{15.81\,{$\pm$}\,3.58} \\
    & JUF-MRI (LOUPE) & \underline{29.57\,{$\pm$}\,2.32} & \underline{0.8176\,{$\pm$}\,0.0278} & \underline{8.76\,{$\pm$}\,2.24} & \underline{27.41\,{$\pm$}\,2.27} & \underline{0.6826\,{$\pm$}\,0.0537} & \underline{11.23\,{$\pm$}\,2.81} & \underline{27.00\,{$\pm$}\,2.26}& \underline{0.6447\,{$\pm$}\,0.0609} & \underline{11.76\,{$\pm$}\,2.93} & 23.77\,{$\pm$}\,1.80 & 0.4739\,{$\pm$}\,0.0782 & 16.87\,{$\pm$}\,3.58 \\
    & \textbf{JUF-MRI (Ours)} & \textbf{30.24\,{$\pm$}\,2.28} & \textbf{0.8526\,{$\pm$}\,0.0243} & \textbf{8.10\,{$\pm$}\,2.04} & \textbf{27.57\,{$\pm$}\,2.25} & \textbf{0.6993\,{$\pm$}\,0.0531} & \textbf{11.02\,{$\pm$}\,2.74} & \textbf{27.09\,{$\pm$}\,2.24} & \textbf{0.6574\,{$\pm$}\,0.0610} & \textbf{11.64\,{$\pm$}\,2.88} & \textbf{25.36\,{$\pm$}\,2.12} & \textbf{0.5048\,{$\pm$}\,0.0839} & \textbf{14.15\,{$\pm$}\,3.35} \\
    \bottomrule
    \end{tabular}
}
\end{sidewaystable*}

\subsection{Experiments on MC-MRI}
To comprehensively assess the robustness of our JUF-MRI method, we performed reconstruction experiments across multiple modalities using different reference images. Our evaluation compares JUF-MRI against several classic and SOTA deep learning methods, including Zero-Filled (ZF) as baseline, UNet~\cite{ronneberger2015u}, Multi-UNet, CUNet~\cite{deng2020deep}, MDUNet~\cite{xiang2018deep}, Restormer~\cite{zamir2022restormer}, VANet~\cite{lei2023decomposition}, and MC-DuDoN~\cite{lei2024joint}, all implemented according to their official specifications with default parameters. The ZF baseline applies the inverse Fourier transform directly to zero-filled $k$-space data. For MDUNet and Restormer architectures, we concatenate the reference and under-sampled target images along the channel dimension as network input. For VANet and MC-DuDoN implementations, we maintain the iteration number at $4$ to ensure fair comparison. It should be emphasized that MC-DuDoN represents another approach utilizing joint optimization of under-sampling patterns through learning-based methods. Therefore, we test the results of MC-DuDoN and our JUF-MRI under both equispaced and learned masks.

\subsubsection{Reconstruction T2-weighted with PD-weighted}
As shown in TABLE~\ref{tab:ixi_results}, we present the numerical comparison in the IXI dataset. Obviously, our JUF-MRI consistently outperforms all benchmark methods across various metrics. When employing the 1D equispaced sampling mask, while Restormer and VANet show competitive performance across the four acceleration rates (4$\times$, 8$\times$, 10$\times$, and 30$\times$), both MC-DuDoN and our JUF-MRI significantly surpass the ZF baseline. Notably, JUF-MRI achieves a 0.31 dB PSNR improvement over MC-DuDoN at 4$\times$ acceleration, with this performance gap widening to 0.61 dB at 8$\times$ acceleration. The superiority of our method is maintained at higher acceleration rates (10$\times$ and 30$\times$), demonstrating its robust reconstruction capability across varying sampling densities.

For the learned mask, MC-DuDoN employs the LOUPE scheme, achieving an average PSNR improvement of $5.86$ dB across various acceleration rates. We also implement the LOUPE scheme within our JUF-MRI to comprehensively evaluate its advantages. The results show that our JUF-MRI (LOUPE) variant exhibits consistent outperformance over MC-DuDoN (LOUPE) at acceleration ratios of 4$\times$, 8$\times$, and 30$\times$, confirming the robustness of our reconstruction network. Compared to MC-DuDoN (LOUPE) and JUF-MRI (LOUPE), our proposed JUF-MRI combines FEP for mask learning, with an average gain of $6.74$ dB compared to equidistant masks, demonstrating excellent performance. These results demonstrate the efficacy of the learnable mask and also prove the effectiveness of our proposed mask learning strategy guided by the FEP.

\subsubsection{Reconstruction T2-weighted with T1-weighted}
As shown in TABLE~\ref{tab:brats_results}, we provide the performance comparison of these methods in the BraTS2018 dataset. Obviously, our JUF-MRI demonstrates superior reconstruction performance compared to benchmark methods in the majority of test cases. Consistent with the PD-weighted reference results, the proposed framework significantly outperforms the ZF baseline for both equispaced and learned sampling masks across all acceleration rates. While yielding second-best performance at 8$\times$ and 10$\times$ acceleration ratios under the equispaced mask, JUF-MRI achieves optimal reconstruction quality at both lower (4$\times$) and higher (30$\times$) acceleration regimes, demonstrating robust performance across varying sampling densities. This further demonstrates the efficacy and superiority of the proposed joint optimization strategy used in our JUF-MRI framework.

\subsubsection{Reconstruction FSPD-weighted with PD-weighted}
As shown in TABLE~\ref{tab:fastmri_results}, we provide the performance comparison of these methods in the fastMRI dataset. As presented in the table, despite reconstructing images of FSPD-weighted modality being more challenging, our JUF-MRI still achieves the best reconstruction performance across all benchmarks. Despite the inherent challenge of reconstructing FSPD-weighted modality images, JUF-MRI achieves the best performance across all benchmarks. Notably, JUF-MRI demonstrates a $0.34$ dB PSNR advantage over MC-DuDoN at the challenging 30$\times$ acceleration regime. More significantly, our method maintains this performance superiority consistently across higher acceleration ratios, establishing its robustness in extreme under-sampling conditions.

\subsubsection{Visual Comparison}
\begin{figure*}[htbp]
    \centering
    \includegraphics[width=1\textwidth]{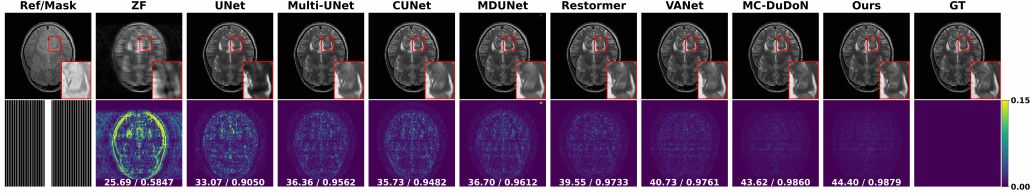}\vspace{-0.1in}
    \caption{Visual comparison under 1D equispaced under-sampling mask on the IXI dataset (4$\times$ acceleration). The first row presents the reconstruction results obtained by different methods, while the second row displays the corresponding error maps relative to the GT.}
    \label{IXI_x4}
\end{figure*}

In addition to the quantitative results, we also report the visual quality. The reconstruction and error maps on the IXI data set with an acceleration of 4$\times$ with a 1D equispaced mask are presented in Fig.~\ref{IXI_x4}. With the PD-weighted reference (first row, first column), the T1-weighted modality images were reconstructed under a 1D equispaced mask (second row, first column). For better visualization, we zoom in on the distinguished region located in the bottom right corner of the reconstructed image in the first row and the corresponding error map in the second row of Fig.~\ref{IXI_x4}. From the zoomed-in view, it can be observed that the image reconstructed by our method exhibits clearer boundaries and enhanced contrast between gray matter and white matter compared to other methods. Furthermore, the error maps reveal that our method produces the lowest overall discrepancy from the ground truth, demonstrating its superior reconstruction performance.

\begin{figure*}[htbp]
    \centering
    \includegraphics[width=1\textwidth]{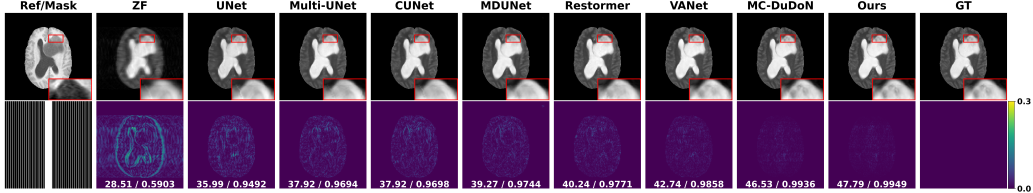}\vspace{-0.1in}
    \caption{Visual comparison under 1D equispaced under-sampling mask on the Brats2018 dataset (4$\times$ acceleration). The first row presents the reconstruction results obtained by different methods, while the second row displays the corresponding error maps relative to the GT.}
    \label{Brats2018_x4}
\end{figure*}

Fig.~\ref{Brats2018_x4} presents qualitative results following the same layout as the IXI dataset.
The zoomed-in view illustrates that our method reconstructed the necrotic core of glioblastoma, which is the dark area within the high signal lesion in the upper right quadrant, with higher structural clarity and contrast compared to other methods. In addition, the error plot indicates that our method achieves lower overall reconstruction errors while more effectively preserving the overall anatomical structure.

\begin{figure*}[htbp]
    \centering
    \includegraphics[width=1\textwidth]{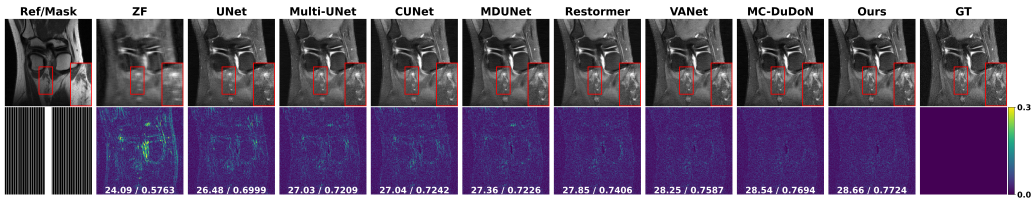}\vspace{-0.1in}
    \caption{Visual comparison under 1D equispaced under-sampling mask on the FastMRI dataset (4$\times$ acceleration). The first row presents the reconstruction results obtained by different methods, while the second row displays the corresponding error maps relative to the GT.}
    \label{fastmri_x4}
\end{figure*}

Fig.~\ref{fastmri_x4} presents qualitative comparisons under 4$\times$ acceleration with a 1D equispaced mask. As shown in the zoomed-in ROI, JUF-MRI yields more accurate edge delineation and better recovery of fine anatomical details in knee MR images. Furthermore, the lower reconstruction errors shown in the error maps highlight the model's ability to preserve structural integrity and its robust generalization across diverse anatomical structures.

\subsection{Complexity and Efficiency Analysis}
In this section, we analyze the computational complexity and practical efficiency of all compared methods. 
The results, summarized in Table \ref{Complexity} ($10\times$ acceleration, IXI dataset), highlight a critical trade-off among model size (parameters), theoretical computational cost (FLOPs), and real-world runtime performance.

A primary finding is the exceptional parameter efficiency of our proposed JUF-MRI framework. Compared with SOTA methods such as Restormer and MDUNet, which employ 26.10 M and 10.36 M parameters, respectively, our JUF-MRI achieves superior reconstruction quality (40.47 dB PSNR) using only 2.11 M parameters, corresponding to a more than 12-fold reduction in model size relative to Restormer.

This parameter efficiency is enabled by our model-driven deep unfolding architecture, which reuses a compact network block (2.11 M parameters) across multiple iterative stages. While this iterative design incurs a relatively higher theoretical computational cost (483.67 GFLOPs), it allows for substantial model compression without sacrificing reconstruction quality.

However, the high theoretical FLOPs of our framework do not translate into impractical inference times. The actual runtime of JUF-MRI is 106.59 ms, which is highly competitive and slightly faster than Restormer (110.96 ms), despite the latter being $12\times$ larger. This efficiency advantage can be attributed to the substantial memory footprint of Restormer (26.10 M parameters), which induces a memory-bandwidth bottleneck, whereas the compact 2.11 M parameter model of JUF-MRI allows for more efficient GPU execution.

\begin{table}[htbp]
    \centering
    \renewcommand{\arraystretch}{1.2}
    \caption{Quantitative Comparison of Model Complexity (FLOPs, Parameters, Runtime) and Reconstruction Quality Metrics ($10\times$ Acceleration on the IXI Dataset).}
    \resizebox{\columnwidth}{!}{
    \begin{tabular}{l|ccccc}
        \toprule
        \textbf{Method} & \textbf{FLOPs (G)} & \textbf{Params (M)} & \textbf{Runtime (ms)} & \textbf{PSNR (dB)$\uparrow$} & \textbf{SSIM$\downarrow$} \\
        \midrule
        UNet\cite{ronneberger2015u}          & 9.56  & 0.26  & 3.10  & 27.38\,{$\pm$}\,2.14 & 0.8255\,{$\pm$}\,0.0366 \\
        Multi-UNet                          & 10.77 & 0.28  & 3.10  & 35.33\,{$\pm$}\,2.22 & 0.9562\,{$\pm$}\,0.0149 \\
        CUNet\cite{deng2020deep}            & 14.27 & 0.22  & 41.26 & 34.94\,{$\pm$}\,2.02 & 0.9498\,{$\pm$}\,0.0157 \\
        MDUNet\cite{xiang2018deep}          & 116.47 & 10.36 & 9.87  & 37.39\,{$\pm$}\,2.33 & 0.9650\,{$\pm$}\,0.0135 \\
        Restormer\cite{zamir2022restormer}  & 141.15 & 26.10 & 110.96 & 38.42\,{$\pm$}\,2.51 & 0.9708\,{$\pm$}\,0.0114 \\
        VANet\cite{lei2023decomposition}    & 139.86 & 1.39  & 38.06 & 38.16\,{$\pm$}\,2.44 & 0.9657\,{$\pm$}\,0.0119 \\
        MC-DuDoN\cite{lei2024joint}         & 294.18 & 1.46  & 64.95 & \underline{40.09\,{$\pm$}\,2.60} & \underline{0.9757\,{$\pm$}\,0.0094} \\
        \textbf{JUF-MRI (Ours)}             & \textbf{483.67} & \textbf{2.11} & \textbf{106.59} & \textbf{40.47\,{$\pm$}\,2.63} & \textbf{0.9778\,{$\pm$}\,0.0091} \\
        \bottomrule
    \end{tabular}
    }
    \label{Complexity}
\end{table}

In conclusion, JUF-MRI achieves an excellent balance between reconstruction quality, model compactness, and computational efficiency. It attains SOTA reconstruction performance while maintaining a minimal parameter count and practical inference runtime comparable to more complex models. This combination of high accuracy and low memory footprint makes JUF-MRI particularly well-suited for deployment in real-world clinical systems, where model size and memory constraints are critical considerations.

\subsection{Statistical Significance Analysis}
\begin{table}[htbp]
\centering
\renewcommand{\arraystretch}{1.4} 
\caption{The methods achieving the highest reconstruction metrics at each acceleration rate across the three datasets and the results of paired t-tests.}
\resizebox{\columnwidth}{!}{
\begin{tabular}{l|l|rrrr}
\toprule
Dataset & Mask & 4$\times$Acceleration & 8$\times$Acceleration & 10$\times$Acceleration & 30$\times$Acceleration \\
\midrule
\multirow{2}{*}{IXI} & Equispaced & JUF-MRI\stat & JUF-MRI\stat & JUF-MRI\stat & JUF-MRI\stat \\
                      & Learned    & JUF-MRI(FEP)\stat & JUF-MRI(FEP)\stat & JUF-MRI(FEP)\stat & JUF-MRI(FEP)\stat \\
\midrule
\multirow{2}{*}{Brats2018} & Equispaced & JUF-MRI\stat & MC-DuDoN\stat & MC-DuDoN\stat & JUF-MRI\stat \\
                            & Learned    & JUF-MRI(FEP)\stat & JUF-MRI(FEP)$^*$ & JUF-MRI(FEP)\stat & JUF-MRI(FEP)\stat \\
\midrule
\multirow{2}{*}{FastMRI} & Equispaced & JUF-MRI\stat & JUF-MRI\stat & JUF-MRI\stat & JUF-MRI$^*$ \\
                          & Learned    & JUF-MRI(FEP)\stat & JUF-MRI(FEP)\stat & JUF-MRI(FEP)\stat & JUF-MRI(FEP)\stat \\
\bottomrule
\end{tabular}}
\label{t-test}
\end{table}

To rigorously validate our quantitative results and ensure that the observed performance gains were not due to random variations, we performed paired t-tests comparing our proposed method, JUF-MRI (Ours), with other methods on the three datasets across different acceleration rates using PSNR, SSIM, and RMSE as evaluation metrics. Table \ref{t-test} presents the methods achieving the best reconstruction metrics under different conditions along with the results of paired t-tests. In the table, $^{***}$ indicates that under the same conditions (dataset, mask, and acceleration rate), the method outperforms all other reconstruction methods with a highly significant difference ($p < 0.001$), while $^*$ indicates that the method outperforms all other reconstruction methods with a statistically significant difference ($p < 0.05$). From the results in the table, we can observe that in most cases, JUF-MRI outperforms other reconstruction methods with highly significant improvements in reconstruction metrics. In a few cases (the learned mask at $8\times$ acceleration on the Brats2018 dataset and the equispaced mask at $30\times$ acceleration on the FastMRI dataset), JUF-MRI still achieves superior reconstruction performance with statistically significant results. The only exceptions occur under the equispaced masks at $8\times$ and $10\times$ acceleration on the Brats2018 dataset, where its reconstruction metrics are slightly lower than those of MC-DuDoN.

\section{Ablation Study}
\subsection{Effectiveness of FEP}
\begin{figure}[htbp]
    \centering

    \subfloat[\normalsize IXI]{%
        \includegraphics[width=0.32\linewidth]{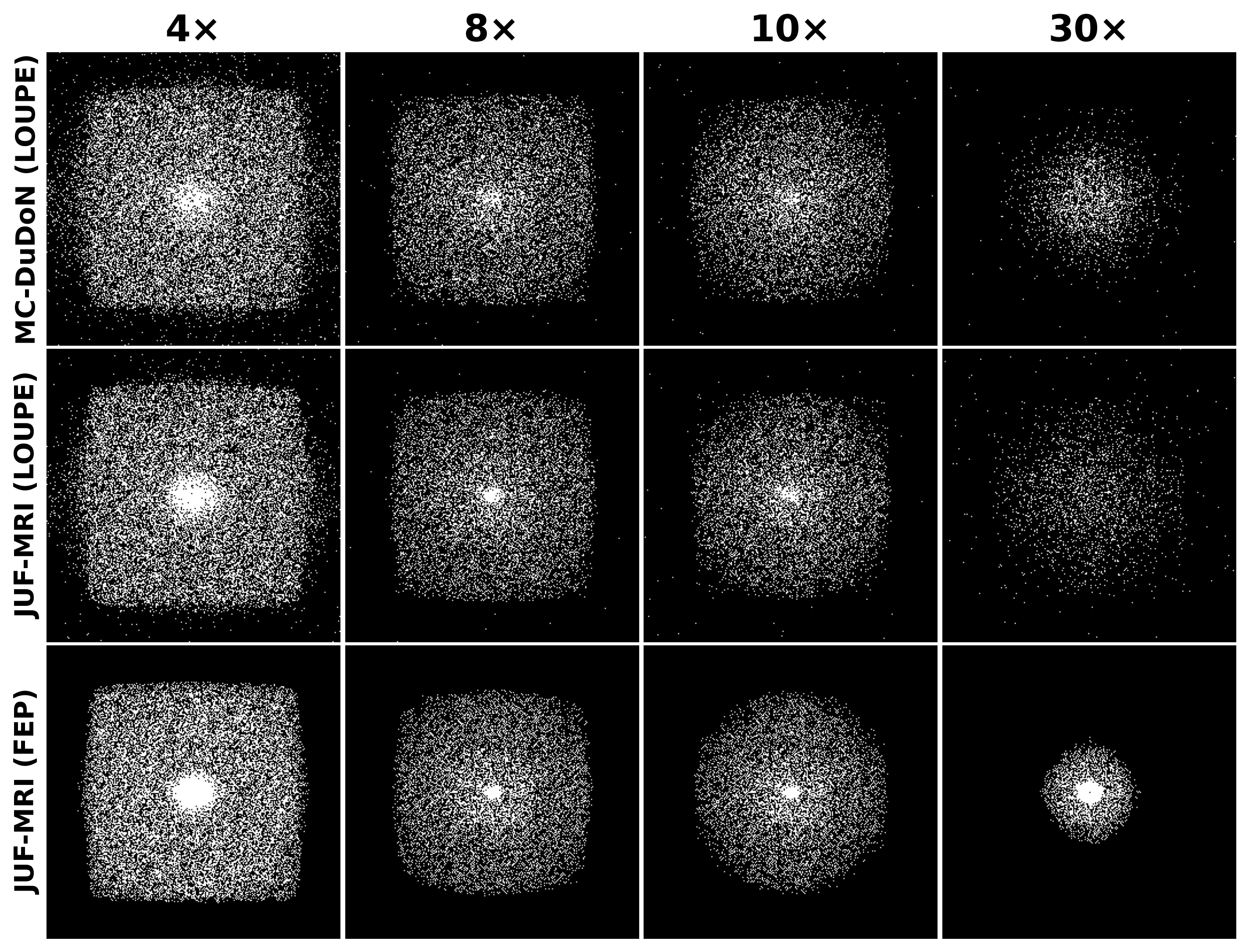}%
        \label{fig:mask_ixi}}%
    \hfill 
    \subfloat[\normalsize BraTS2018]{%
        \includegraphics[width=0.32\linewidth]{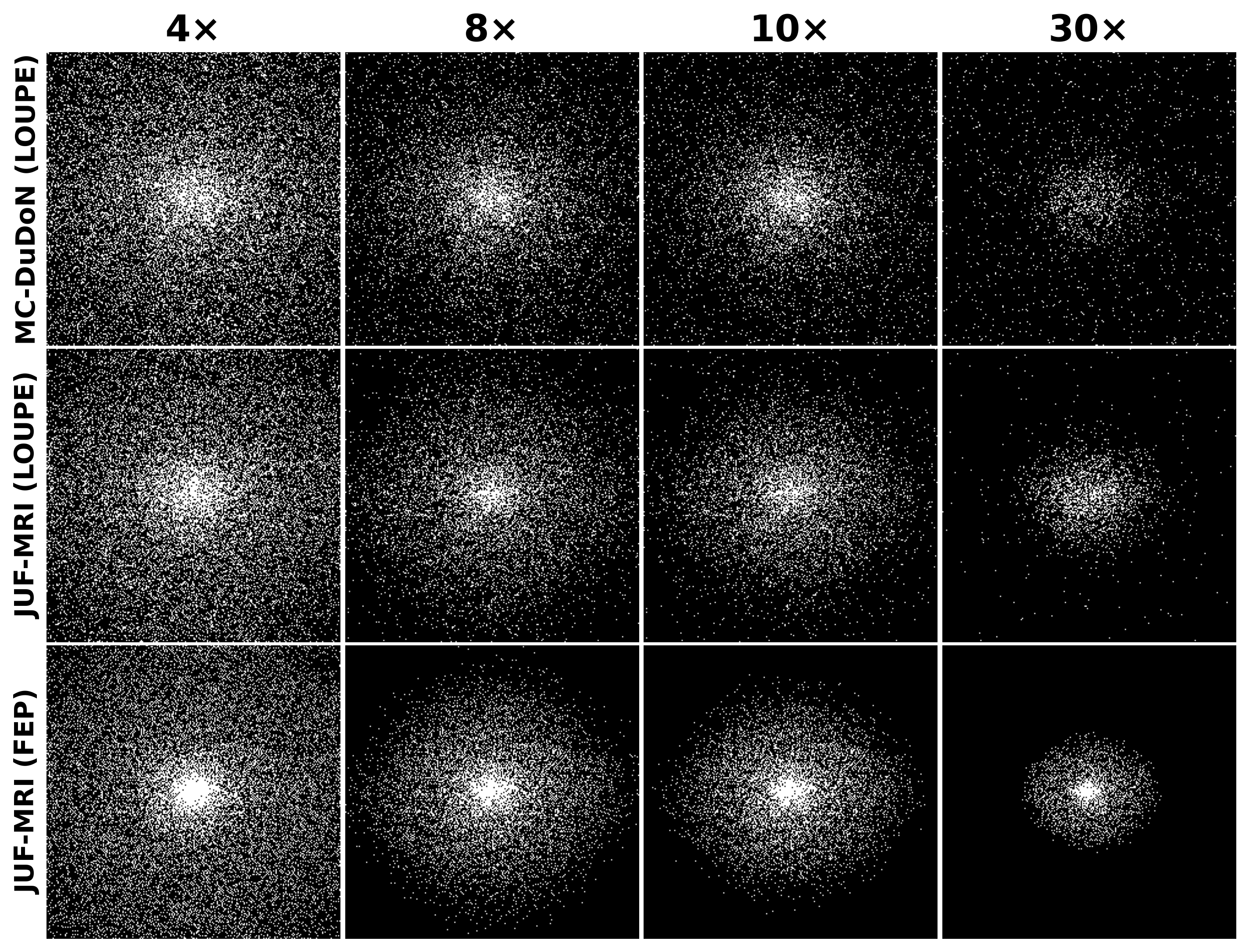}%
        \label{fig:mask_brats}}%
    \hfill
    \subfloat[\normalsize FastMRI]{%
        \includegraphics[width=0.32\linewidth]{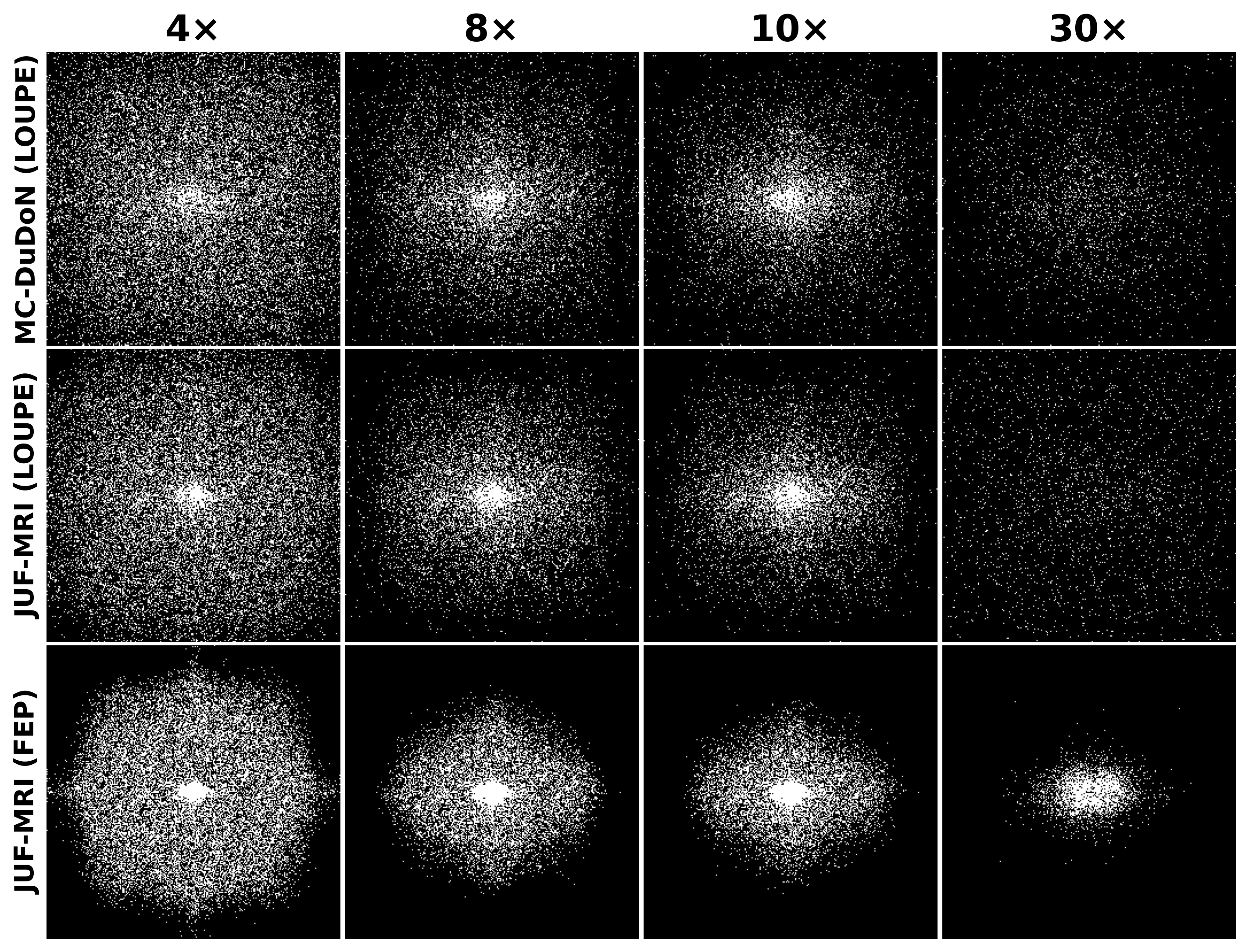}%
        \label{fig:mask_fastmri}}

    \caption{Under-sampling masks jointly optimized on different datasets and acceleration ratios. The first row shows the masks jointly optimized by MC-DuDoN and LOUPE frameworks. The second row shows the masks jointly optimized by JUF-MRI and LOUPE. The third row shows the masks learned by our method. Please zoom in to see the details.}
    \label{fig:learned_masks_all}
\end{figure}

\begin{figure}[htbp]
    \centering
    \includegraphics[width=0.8\textwidth]{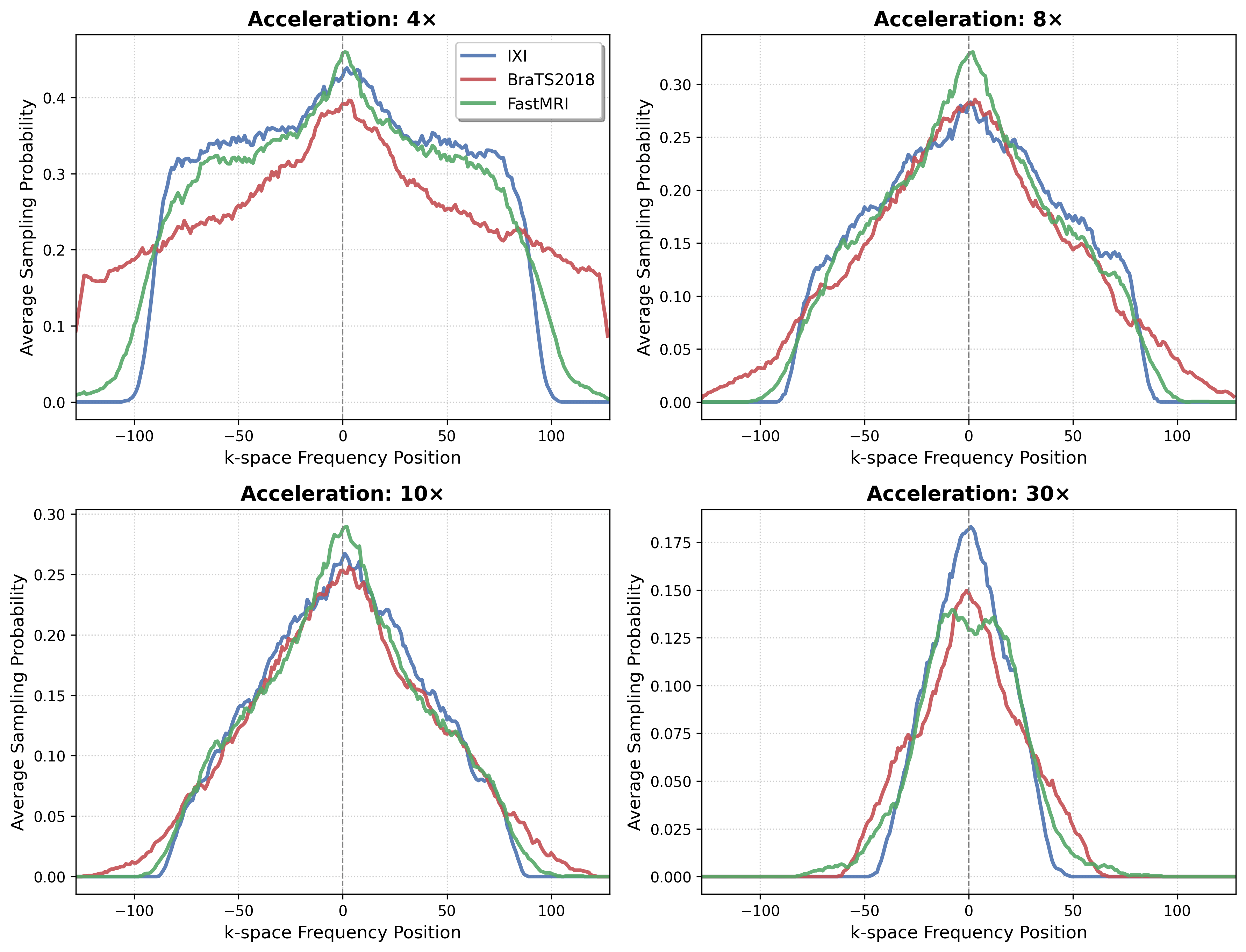}
    \caption{1D Sampling density profiles of learned masks across three datasets ($4\times, 8\times, 10\times, 30\times$).}
    \label{fig: 1D Sampling Density}
\end{figure}

In this work, the FEP is introduced to more effectively utilize complementary information from reference images, with its inclusion during optimization producing markedly different sampling patterns, as demonstrated in Fig.~\ref{fig:learned_masks_all}. This figure systematically compares jointly optimized under-sampling masks across multiple acceleration ratios and datasets under three configurations: (1) MC-DuDoN with LOUPE (first row), (2) JUF-MRI with LOUPE (second row), and (3) JUF-MRI incorporating the FEP (third row). The results demonstrate that integrating FEP produces sampling masks with significantly higher density in the central $k$-space region. In contrast to LOUPE-based methods, the masks exhibit a frequency-aware structure, with dense sampling of low-frequency components and gradually decreasing coverage toward higher frequencies. This reveals a more effective modeling of frequency-dependent importance across the spectrum. Therefore, the proposed method achieves excellent reconstruction results, outperforming LOUPE-based strategies that lack such frequency informed guidance. Although the learned masks incorporating the FEP show some variations across different datasets due to differences in data distribution, they generally exhibit similar sampling patterns. As shown in Fig. \ref{fig: 1D Sampling Density}, the learned masks incorporating the FEP for all three datasets share a common characteristic — higher sampling density in the low-frequency region, especially near the center, and progressively lower sampling probability toward the outer regions.

\begin{figure}[htbp]
    \centering
    \includegraphics[width=0.88\linewidth]{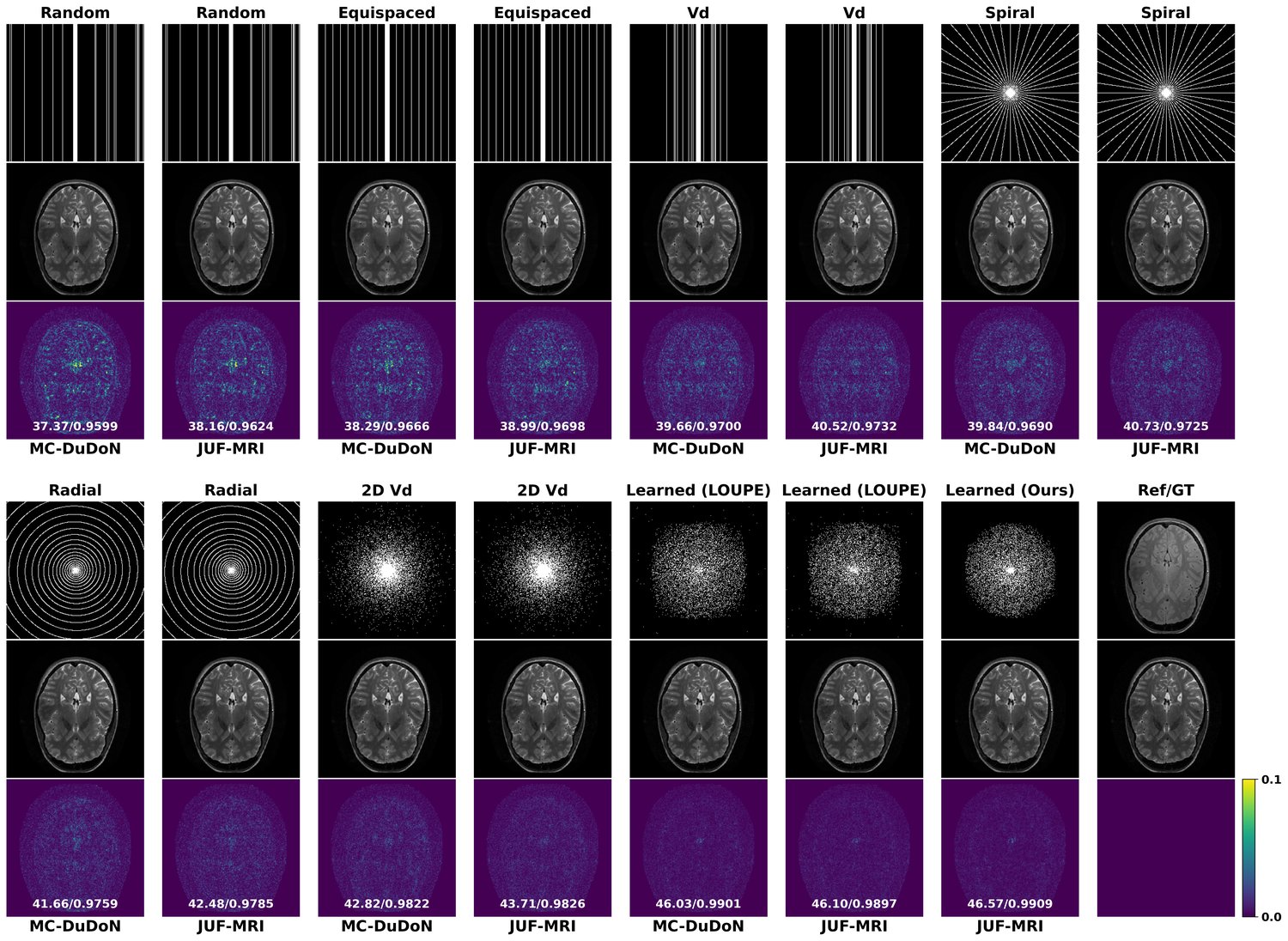}\\[2mm] 
    \includegraphics[width=0.88\linewidth]{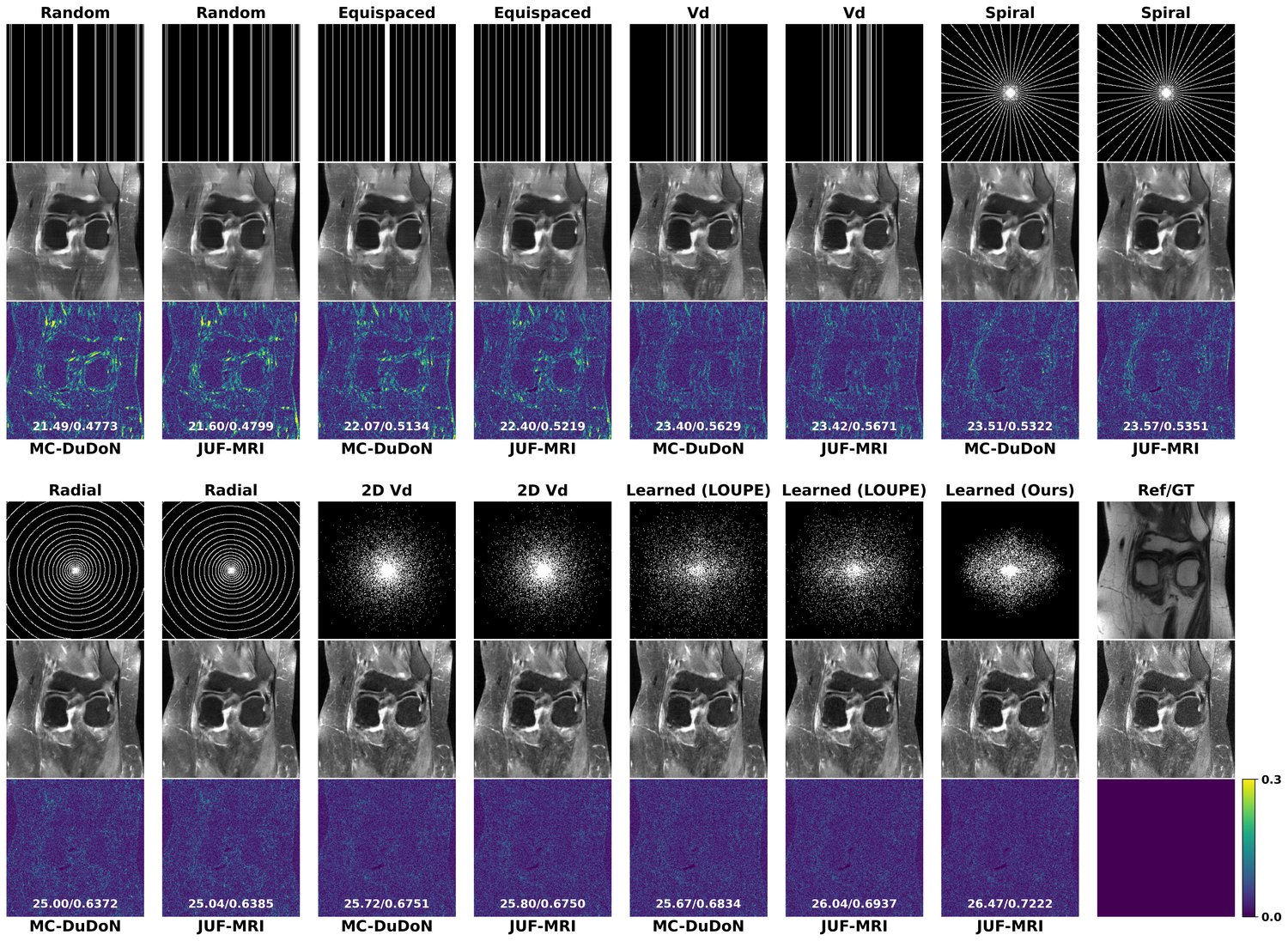}
    \caption{Visual comparison using MC-DuDoN and JUF-MRI as the reconstruction framework under 10$\times$ acceleration with different under-sampling masks. Top: IXI dataset. Bottom: FastMRI dataset. Vd denotes variable-density mask, and 2D Vd represents two-dimensional variable-density mask.}
    \label{fig:mask_contrast_two_row}
\end{figure}

As shown in TABLES~\ref{tab:ixi_results}, ~\ref{tab:brats_results}, and~\ref{tab:fastmri_results}, our method JUF-MRI achieves excellent reconstruction performance across multiple datasets and acceleration ratios. To further validate the superiority of the masks learned through joint optimization with our proposed FEP, we conducted additional experiments on the IXI and FastMRI datasets, using JUF-MRI as the reconstruction backbone and comparing against different under-sampling masks (10$\times$ acceleration). Fig.~\ref{fig:mask_contrast_two_row} presents comparative results across different under-sampling masks.
The first row shows various masks.
Subsequent rows show reconstructed images from MC-DuDoN and JUF-MRI with corresponding error maps. The proposed FEP-optimized mask consistently achieves superior reconstruction quality when paired with JUF-MRI, demonstrating that our frequency domain prior effectively leverages cross-modality structural information to improve reconstruction quality.

\subsection{Impact of the Number of Iterative Sub-modules}
\begin{figure}[htbp]
    \centering
    \includegraphics[width=0.8\textwidth]{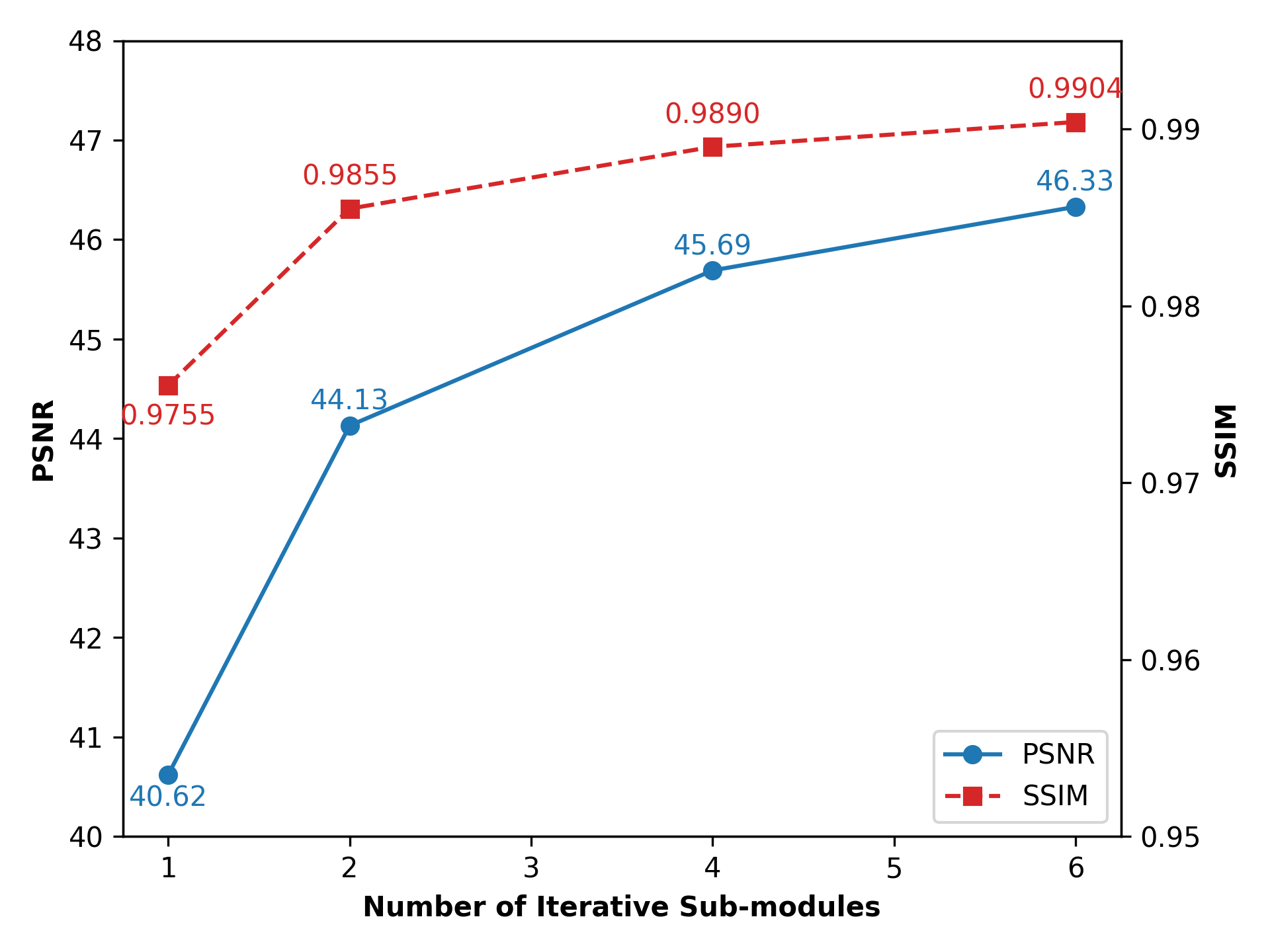}
    \caption{Performance comparison under different numbers of iterative sub-modules in the proposed JUF-MRI.}
    \label{fig: iteration_effect}
\end{figure}
The number of iterative sub-modules in our deep unfolding network corresponds to the unfolding steps, emulating the iterative refinement of an optimization algorithm. 
To investigate this relationship, we evaluated configurations with 1, 2, 4, and 6 sub-modules on the IXI dataset using a 1D equispaced mask at 4$\times$ acceleration. As shown in Fig.~\ref{fig: iteration_effect}, increasing the number of iteration modules can improve the reconstruction performance. However, the improvement decreases with the number of iterations. In addition, deeper unfolded networks lead to slower convergence, increased memory usage during training, and longer inference time. Therefore, $4$ iterator modules are adopted in the final model to strike a balance between reconstruction quality and computational efficiency.

\subsection{Effect of Proposed Strategies}

\begin{table*}[t]
\centering
\caption{Ablation studies on IXI, Brats2018, and FastMRI datasets under a 1D equispaced 4$\times$ acceleration mask. (RM stands for the reference modality, DS stands for the decomposition strategy，and FDL stands for the frequency-domain loss)}
\resizebox{\textwidth}{!}{
\begin{tabular}{l|l|rrrrr}
\toprule
\textbf{Datasets} & \textbf{Metrics} & \textbf{w/o RM} & \textbf{w/o $k$-space data} & \textbf{w/o DS} & \textbf{w/o FDL} & \textbf{Full Model} \\ \midrule
\multirow{3}{*}{IXI} & PSNR$\uparrow$ & 41.41 $\pm$ 2.54 & 45.76 $\pm$ 2.68 & 44.79 $\pm$ 2.62 & 45.62 $\pm$ 2.66 & \textbf{45.89 $\pm$ 2.71} \\
 & SSIM$\uparrow$ & 0.9809 $\pm$ 0.0065 & 0.9907 $\pm$ 0.0037 & 0.9884 $\pm$ 0.0045 & 0.9900 $\pm$ 0.0039 & \textbf{0.9905 $\pm$ 0.0037} \\
 & RMSE$\downarrow$ & 2.26 $\pm$ 0.63 & 1.37 $\pm$ 0.39 & 1.53 $\pm$ 0.43 & 1.39 $\pm$ 0.40 & \textbf{1.36 $\pm$ 0.39} \\ \midrule
\multirow{3}{*}{Brats2018} & PSNR$\uparrow$ & 43.81 $\pm$ 2.17 & 45.97 $\pm$ 2.38 & 44.33 $\pm$ 2.20 & 45.55 $\pm$ 2.27 & \textbf{46.05 $\pm$ 2.34} \\
 & SSIM$\uparrow$ & 0.9911 $\pm$ 0.0047 & 0.9941 $\pm$ 0.0032 & 0.9920 $\pm$ 0.0026 & 0.9939 $\pm$ 0.0022 & \textbf{0.9945 $\pm$ 0.0021} \\
 & RMSE$\downarrow$ & 1.70 $\pm$ 0.43 & 1.33 $\pm$ 0.37 & 1.60 $\pm$ 0.40 & 1.39 $\pm$ 0.36 & \textbf{1.32 $\pm$ 0.35} \\ \midrule
\multirow{3}{*}{FastMRI} & PSNR$\uparrow$ & 27.84 $\pm$ 2.28 & 28.19 $\pm$ 2.26 & 28.05 $\pm$ 2.25 & 28.18 $\pm$ 2.25 & \textbf{28.19 $\pm$ 2.26} \\
 & SSIM$\uparrow$ & 0.7428 $\pm$ 0.0367 & 0.7556 $\pm$ 0.0364 & 0.7495 $\pm$ 0.0364 & 0.7556 $\pm$ 0.0362 & \textbf{0.7557 $\pm$ 0.0362} \\
 & RMSE$\downarrow$ & 10.68 $\pm$ 2.70 & 10.26 $\pm$ 2.57 & 10.42 $\pm$ 2.60 & 10.27 $\pm$ 2.56 & \textbf{10.26 $\pm$ 2.57} \\ \bottomrule
\end{tabular}
}
\label{tab:ablation_study}
\end{table*}

In this part, we evaluate the impact of incorporating reference modality information, $k$-space data, and decomposition strategy on the performance of the reconstruction model. Since our proposed JUF-MRI is based on a deep unfolding architecture, removing these variables will lead to corresponding changes in the network structure and objective function. Therefore, the objective function can be reformulated to 
\begin{equation}
\small
\begin{aligned}
\mathop{\arg\min}\limits_{\left \{X,K  \right \}} \ & 
\frac{1}{2}\left \| M\mathcal{F}(X)-\widetilde{K}\right \|_{F}^{2}
+\frac{\alpha }{2}\left \|K - \mathcal{F}(X) \right \|_{F}^{2} 
\\&+ \lambda_{1} \mathcal{R}_{1}(X)
+ \lambda_{2} \mathcal{R}_{2}(K),
\end{aligned}
\label{objective function without reference}
\end{equation}
\begin{equation}
\small
\begin{aligned}
\mathop{\arg\min}\limits_{\left \{X,S,D,\phi  \right \}} \ & 
\frac{1}{2}\left \| M\mathcal{F}(X)-\widetilde{K}\right \|_{F}^{2}
+ \frac{\gamma}{2}\left \| S + D - Y_\text{SA} \right \|_{F}^{2}
\\&
 + \frac{\beta }{2}\left \|AX - BS \right \|_{F}^{2}+ \lambda_{1} \mathcal{R}_{1}(X)
+ \lambda_{2} \psi_{1}(S) \\& + \lambda_{3} \psi_{2}(D)
+ \lambda_{4} \Phi_{1}(\phi),
\end{aligned}
\label{objective function without kspace}
\end{equation}
\begin{equation}
\small
\begin{aligned}
\mathop{\arg\min}\limits_{\left \{X,K,\phi  \right \}} \ & 
\frac{1}{2}\left \| M\mathcal{F}(X)-\widetilde{K}\right \|_{F}^{2}
+ \frac{\alpha }{2}\left \|K - \mathcal{F}(X) \right \|_{F}^{2} 
\\&
 + \frac{\beta }{2}\left \|AX - BY_{SA} \right \|_{F}^{2}+ \lambda_{1} \mathcal{R}_{1}(X)
\\&+ \lambda_{2} \mathcal{R}_{2}(K)  + \lambda_{3} \Phi_{1}(\phi).
\end{aligned}
\label{objective function without decomposition strategy}
\end{equation}
Eq.~\eqref{objective function without reference} represents the objective function without the reference modality, eliminating the need to consider spatial misalignment and data noise. Eq.~\eqref{objective function without kspace} represents the objective function without $k$-space data. Eq.~\eqref{objective function without decomposition strategy} represents the objective function without the use of the decomposition strategy. We unfold these three objective functions into deep networks using the TITAN algorithm~\cite{phan2023inertial}. 
 
The results of the aforementioned three variants are present in TABLE~\ref {tab:ablation_study}. The table lists the ablation study results on three datasets under a 1D equispaced 4$\times$ acceleration mask. The results demonstrate that incorporating the reference image, $k$-space data, and decomposition strategy all contribute positively to the reconstruction performance of the network. Among them, the inclusion of the reference image has the most significant impact, indicating that it indeed contains redundant information beneficial for reconstructing the target modality. The use of a decomposition strategy effectively filters out irrelevant information, thereby enhancing the performance of the reconstruction model. 

\subsection{Effect of the Combined Frequency-domain Loss}

In order to better utilize high-frequency components to improve the recovery of fine structures, we propose a combined frequency-domain loss. The Combined Frequency-domain Loss is controlled by four hyperparameters: $\lambda_{1}$ and $\lambda_{2}$, which are used to control the proportion of the high-frequency and low-frequency components in the Frequency loss, and $\alpha$ and $\beta$, which represent the balance between the Frequency loss and the $\ell_ {1} $ loss, respectively.

We conduct a sensitivity analysis and ablation study to validate the design of our proposed combined frequency-domain loss function. The results are presented in Table \ref{ablation_hyper}. First, in the first five rows of the table, we analyze the contribution of the $\ell_1$ and Frequency loss terms by varying $\alpha$ and $\beta$, while keeping the internal frequency weights fixed ($\lambda_1=0.8, \lambda_2=0.2$). From this result, it can be observed that when the loss only contains the frequency loss, the overall performance of the model experiences a significant decline (44.53 dB PSNR). Among the different weighting schemes (Rows 1-4), we observe that the peak performance (45.89 dB PSNR) is achieved with the combination of $\alpha=0.2$ and $\beta=0.8$. This indicates that while both terms are necessary, assigning a higher weight to the primary $\ell_ {1} $ loss yields the optimal balance. Next, we fixed the optimal outer balance ($\alpha=0.2, \beta=0.8$) and analyzed the internal weights of the Frequency loss, $\lambda_1$ (low-freq) and $\lambda_2$ (high-freq). The optimal result is achieved at $\lambda_1=0.8$ and $\lambda_2=0.2$ (Row 2). This empirically validates our hypothesis that a strong emphasis on the low-frequency component is crucial for overall reconstruction fidelity, while the high-frequency component (weighted at $0.2$) effectively contributes to recovering fine details and structures.

\begin{table}[htbp] 
    \centering 
    \renewcommand{\arraystretch}{1.1} 
    \caption{Quantitative Results under Different Hyperparameter Settings of $\lambda_1$, $\lambda_2$,  $\alpha$, and $\beta$ at $10\times$ Acceleration on the IXI Dataset}
    
    \resizebox{\columnwidth}{!}{ 
    \begin{tabular}{cccc|ccc} 
    \hline 
    \textbf{$\lambda_1$} & \textbf{$\lambda_2$} & \textbf{$\alpha$} &\textbf{$\beta$} & \textbf{PSNR$\uparrow$} & \textbf{SSIM$\uparrow$} & \textbf{RMSE$\downarrow$} \\
    \hline 
    $-$ & $-$ & 0 & 1 & 45.62\,{$\pm$}\,2.66 & 0.9900\,{$\pm$}\,0.0039 & 1.39\,{$\pm$}\,0.40 \\ 
    \textbf{0.8} & \textbf{0.2} & \textbf{0.2} & \textbf{0.8} & \textbf{45.89\,{$\pm$}\,2.71} & \textbf{0.9945\,{$\pm$}\,0.0037} & \textbf{1.36\,{$\pm$}\,0.39} \\  
    0.8 & 0.2 & 0.5 & 0.5 & 45.63\,{$\pm$}\,2.67 & 0.9903\,{$\pm$}\,0.0039 & 1.39\,{$\pm$}\,0.40 \\ 
    0.8 & 0.2 & 0.8 & 0.2 & 45.36\,{$\pm$}\,2.67 & 0.9898\,{$\pm$}\,0.0041 & 1.44\,{$\pm$}\,0.41 \\ 
    0.8 & 0.2 & 1 & 0 & 44.53\,{$\pm$}\,2.56 & 0.9880\,{$\pm$}\,0.0046 & 1.58\,{$\pm$}\,0.43 \\ 
    \hline 
    1 & 0 & 0.2 & 0.8 & 45.17\,{$\pm$}\,2.65 & 0.9893\,{$\pm$}\,0.042 & 1.47\,{$\pm$}\,0.41 \\ 
    0.5 & 0.5 & 0.2 & 0.8 & 45.27\,{$\pm$}\,2.63 & 0.9893\,{$\pm$}\,0.0041 & 1.45\,{$\pm$}\,0.41 \\ 
    0.2 & 0.8 & 0.2 & 0.8 & 45.31\,{$\pm$}\,2.67 & 0.9894\,{$\pm$}\,0.0042 & 1.45\,{$\pm$}\,0.41 \\ 
    0.0 & 1 & 0.2 & 0.8 & 45.39\,{$\pm$}\,2.66 & 0.9896\,{$\pm$}\,0.0041 & 1.43\,{$\pm$}\,0.41 \\ 
    \hline 
    \end{tabular} 
    } 
    
    \label{ablation_hyper} 
\end{table}

\section{Discussion}
Effectively exploiting the complementary information from the reference modality is crucial for achieving high-quality performance in MC-MRI reconstruction task. In our proposed method, the reference modality is leveraged in two primary ways. First, a CDM is employed to get the FEP between the reference and target modalities. The FEP is then incorporated into the joint optimization of both the reconstruction network and the under-sampling mask, enabling the learned sampling mask to better reflect information from the reference modality. Second, the reference modality is directly introduced into the reconstruction objective through a decomposition strategy and a spatial alignment network, which together suppress irrelevant or misaligned components. This facilitates more efficient and effective learning of complementary information from the reference image. It is precisely the effective utilization of the reference modality that enables our proposed framework, JUF-MRI, to achieve superior reconstruction performance. 

Moreover, the joint optimization of the reconstruction network and the under-sampling mask enables the model to reduce sampling in redundant frequency regions and prioritize the acquisition of more informative $k$-space data, thereby improving both sampling efficiency and reconstruction quality. 

Although our JUF-MRI framework demonstrates strong reconstruction performance, some key limitations that need to be addressed. The use of CDM introduces additional training complexity and computational overhead, potentially creating a bottleneck for clinical deployment. Furthermore, while this study primarily leverages FEP for under-sampling optimization, the full potential of FEP remains largely untapped and warrants systematic investigation. Another limitation lies in the exclusive use of single-coil datasets for experimentation, and its generalization ability was not fully verified. To address these challenges, future work will concentrate on three key directions: optimizing CDM to reduce computational burden, developing more comprehensive integration strategies for FEP, and extending the framework to multi-coil scenarios to enhance its clinical applicability.

\section{Conclusion}
In this work, we propose JUF-MRI, a novel framework that effectively addresses four fundamental limitations in existing MC-MRI reconstruction methods: (1) non-physical network architectures, (2) isolated image-domain processing, (3) inefficient reference modality utilization, and (4) fixed equispaced sampling patterns. By integrating the frequency error prior learned from the CDM with the model-driven deep unfolding reconstruction network, JUF-MRI enables joint optimization of both under-sampling strategies and reconstruction. This formulation bridges physical properties of $k$-space with the representational power of deep learning. Meanwhile, the spatial alignment module and reference feature decomposition strategy further enhance cross-modal information transfer. Extensive experiments demonstrate that JUF-MRI consistently outperforms state-of-the-art methods across acceleration factors (4-30$\times$), diverse sampling schemes, and multiple anatomical contrasts, delivering both quantitatively superior results (PSNR/SSIM) and enhanced quality of diagnostic image. This work highlights the potential of JUF-MRI as a physically interpretable scheme by balancing physical interpretability with data-driven performance, offering a robust solution for efficient, high-quality multi-contrast imaging. 

\normalem

\bibliographystyle{elsarticle-num} 
\bibliography{mybib}

\end{document}